\documentclass[pdflatex,sn-mathphys-num,iicol]{sn-jnl}% Math and Physical Sciences Numbered Reference Style
\usepackage{graphicx}%
\usepackage{multirow}%
\usepackage{amsmath,amssymb,amsfonts}%
\usepackage{amsthm}%
\usepackage{mathrsfs}%
\usepackage[title]{appendix}%
\usepackage{xcolor}%
\usepackage{textcomp}%
\usepackage{manyfoot}%
\usepackage{booktabs}%
\usepackage{algorithm}%
\usepackage{algorithmicx}%
\usepackage{algpseudocode}%
\usepackage{listings}%
\usepackage{hyperref}
\usepackage{mathrsfs}
\usepackage[table]{xcolor}
\theoremstyle{thmstyleone}%
\theoremstyle{thmstyletwo}%

\theoremstyle{thmstylethree}%

\begin{document}

\title[Article Title]{FastDDHPose: Towards Unified, Efficient, and Disentangled 3D Human Pose Estimation}

% From Fast3DHPE to DDHPose++: Unified and Efficient Framework for 3D Human Pose Estimation

% From Fast3DHPE to DDHPose++: Toward a Unified and Efficient Framework for 3D Human Pose Estimation


\author[1]{\fnm{Qingyuan} \sur{Cai}}\email{caiqingyuan@mail.bnu.edu.cn}

\author[1]{\fnm{Linxin} \sur{Zhang}}\email{202421081059@mail.bnu.edu.cn}

\author[3]{\fnm{Xuecai} \sur{Hu}}\email{huxc@mail.ustc.edu.cn}

\author[1]{\fnm{Saihui} \sur{Hou}}\email{housaihui@bnu.edu.cn}

\author[1,2]{\fnm{Yongzhen} \sur{Huang}}\email{huangyongzhen@bnu.edu.cn}

\affil[1]{\orgdiv{School of Artificial Intelligence}, \orgname{Beijing Normal University}, \orgaddress{\city{Beijing}, \postcode{100875}, \country{China}}}

\affil[2]{\orgname{Watrix Technology Limited Company Ltd}, \orgaddress{ \city{Beijing}, \postcode{100088}, \country{China}}}

\affil[3]{\orgdiv{AMAP}, \orgname{Alibaba Group  }, \orgaddress{ \city{Beijing}, \postcode{100102}, \country{China}}}

% \affil[3]{\orgdiv{Department}, \orgname{Organization}, \orgaddress{\street{Street}, \city{City}, \postcode{610101}, \state{State}, \country{Country}}}


\abstract{Recent approaches for monocular 3D human pose estimation (3D HPE) have achieved leading performance by directly regressing 3D poses from 2D keypoint sequences. Despite the rapid progress in 3D HPE, existing methods are typically trained and evaluated under disparate frameworks, lacking a unified framework for fair comparison. To address these limitations, we propose Fast3DHPE, a modular framework that facilitates rapid reproduction and flexible development of new methods. By standardizing training and evaluation protocols, Fast3DHPE enables fair comparison across 3D human pose estimation methods while significantly improving training efficiency. Within this framework, we introduce FastDDHPose, a Disentangled Diffusion-based 3D Human Pose Estimation method which leverages the strong latent distribution modeling capability of diffusion models to explicitly model the distributions of bone length and bone direction while avoiding further amplification of hierarchical error accumulation. Moreover, we design an efficient Kinematic-Hierarchical Spatial and Temporal Denoiser that encourages the model to focus on kinematic joint hierarchies while avoiding unnecessary modeling of overly complex joint topologies. Extensive experiments on Human3.6M and MPI-INF-3DHP show that the Fast3DHPE framework enables fair comparison of all methods while significantly improving training efficiency. Within this unified framework, FastDDHPose achieves state-of-the-art performance with strong generalization and robustness in in-the-wild scenarios. The framework and models will be released at: \url{https://github.com/Andyen512/Fast3DHPE}}

\keywords{3D Human Pose Estimation, Diffusion Models, Disentangled Representation, Unified Framework}



\maketitle

\section{Introduction}\label{sec1}
3D Human Pose Estimation (HPE) has potential applications in virtual reality~\cite{hagbi2010shape,cipresso2018past}, human motion understanding
~\cite{zhang2022spatial,lang2025beyond,li2023gait,li2024aerialgait,wang2025ra}, and human-computer interaction~\cite{carroll1997human,jacko2012human}. The goal of 3D HPE is to regress the 3D joints locations of a human in the 3D space using the input of RGB images or 2D pose sequence. Most existing methods adopt a two-stage pipeline, which first predicts 2D joint locations using off-the-shelf estimators~\cite{wang2020deep,chen2018cascaded,cao2017realtime,fang2017rmpe,wang2023decenternet,jiang2023rtmpose}, and then performs 2D-to-3D lifting to obtain the final estimation results.

\begin{figure}[h]
    \centering
    \includegraphics[width=1\columnwidth]{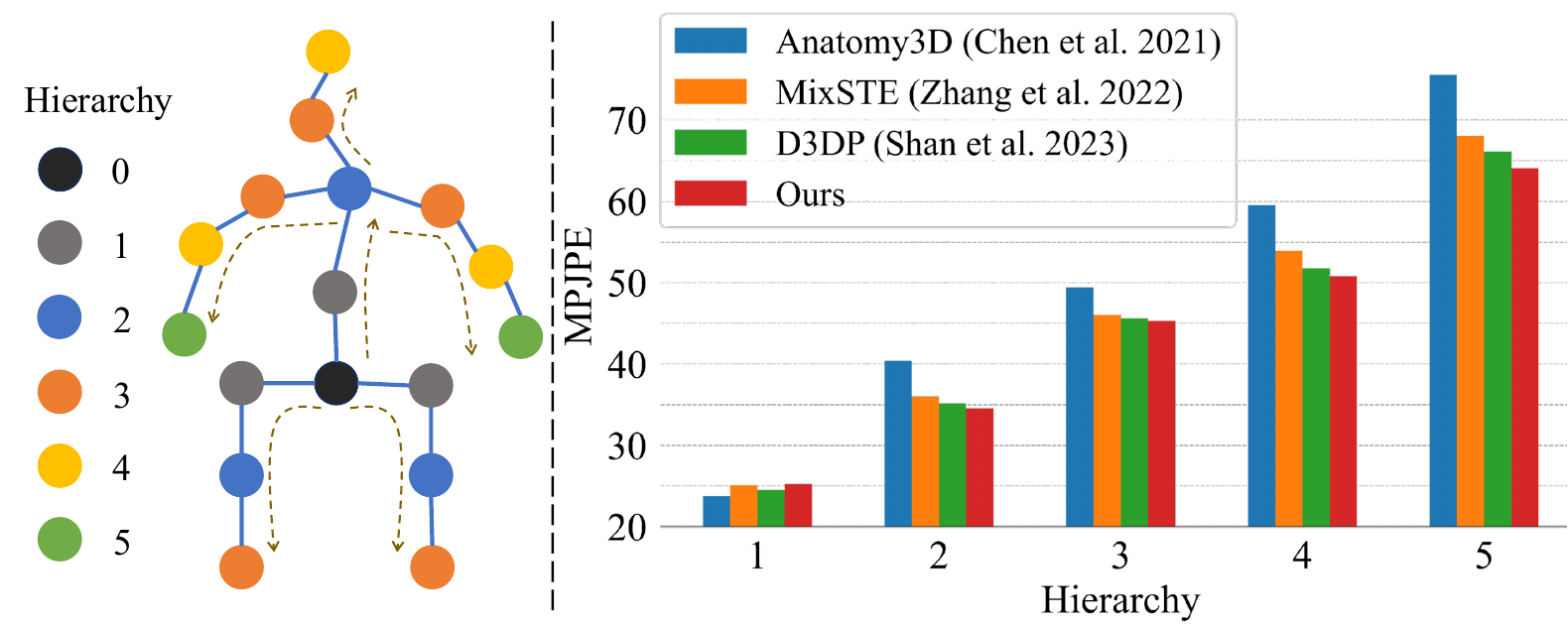} % Reduce the figure size so that it is slightly narrower than the column. Don't use precise values for figure width.This setup will avoid overfull boxes.
    \caption{Left: The hierarchy defined in our method and the forward kinematic structure (drawn with brown dashed lines) based on the Human3.6M dataset. Right: The MPJPE of the hierarchy 1-5 joints comparison among Anatomy3D ~\cite{chen2021anatomy}, MixSTE ~\cite{Zhang_2022_CVPR}, D3DP ~\cite{shan2023diffusion} and our method.}
    % \vspace{-3.85ex}
    \label{fig1}
    \end{figure}
Recently, monocular 3D human pose estimation has experienced significant advancements. Many methods~\cite{liu2025tcpformer,zhang2025pose,zhang2024geometry,yin2025sequential,wang2024uncertainty} have been proposed to alleviate the depth ambiguity. Recently, monocular 3D human pose estimation has achieved significant progress. To alleviate the inherent depth ambiguity in monocular settings, existing methods mainly explore three complementary directions: leveraging temporal context, incorporating spatial--temporal modeling, and introducing explicit human pose priors.

Early representative work such as VideoPose3D~\cite{pavllo:videopose3d:2019} alleviates depth ambiguity by exploiting local temporal consistency through convolutional networks. Building upon this idea, transformer-based methods~\cite{zheng20213d, Zhang_2022_CVPR} further model global spatial–temporal dependencies to better compensate for information loss in the 2D-to-3D lifting process.

Beyond temporal modeling, another line of research mitigates depth ambiguity by explicitly learning or introducing human pose priors. Diffusion-based approaches~\cite{shan2023diffusion, ci2023gfpose, gong2023diffpose, peng2024ktpformer, xu2024finepose} incorporate pose distribution priors during training and formulate 2D-to-3D lifting as a denoising process from noisy pose distributions, thereby improving robustness under uncertain depth conditions. In addition, disentanglement-based methods~\cite{xu2020deep, chen2021anatomy, wang2022motion} explicitly decompose 3D pose estimation into bone length and bone direction prediction, and reconstruct joint locations through forward kinematics of the human skeleton. Such methods introduce explicit structural constraints, including symmetry regularization, joint angle limits~\cite{xu2020deep}, and temporal consistency of bone lengths~\cite{chen2021anatomy}, which further reduce ambiguity and improve physical plausibility.

Despite the significant progress achieved by these methods, existing 3D HPE approaches are often trained and evaluated under different frameworks, which makes direct and fair comparison across methods difficult. Moreover, inefficient training pipelines further limit the scalability and practicality of existing 3D HPE methods. To further advance 3D human pose estimation and better realize its potential, we address this challenge from two complementary perspectives.

% \emph{Revisiting existing 3D HPE methods}: 
% \emph{Exploring a unified and efficient framework}: 

\emph{Reviewing the Past}: 
We make great effort to build a comprehensive open-source framework, \textbf{Fast3DHPE}. 
Within this framework, we systematically review existing 3D human pose estimation methods from the perspectives of data processing, model design, training strategies, and evaluation protocols, and explicitly highlight the key differences among these methods. By integrating diverse methods into a unified framework with a modular design and a rich model zoo, Fast3DHPE provides a consistent and fair benchmarking platform for reliable comparison across existing 3D HPE methods. Moreover, by incorporating Distributed Data Parallel (DDP) and Automatic Mixed Precision (AMP), Fast3DHPE significantly improves training efficiency, leading to substantial speedups across all evaluated methods.

\emph{Advancing the Future}: 
Fast3DHPE provides an integrated framework that stabilizes the training process and regulates experimental settings, enabling more reliable, comparable, and reasonable evaluation across diverse 3D HPE methods. However, there are four problems existing in these methods: \textbf{(1) Limitations of Pose-Space Diffusion for Prior Learning.} Diffusion-based 3D HPE methods~\cite{shan2023diffusion,ci2023gfpose,gong2023diffpose} directly add noise to the original 3D pose which is not conducive to learn the explicit human pose priors such as bone length and bone direction. What if we disentangle the diffusion model by adding noise to bone length and direction separately? This disentangle-based model can separately focus on the temporal consistency of bone length and joint angle variations, better enabling the diffusion model to learn human pose prior. \textbf{(2) Conflict Between Disentangled Priors and Error Accumulation.} Despite the advantages of disentangle-based techniques in incorporating human pose priors, they tend to amplify hierarchical error accumulation during 3D pose estimation. This is because multiple disentangled components, such as bone length and bone direction, propagate their estimation errors through the tree-structured skeleton.  \textbf{(3) Insufficient Modeling of Hierarchical Joint Dependencies.} Although the transformer-based methods have the ability to explore the spatial-temporal context information, these models generally lack attention to the fine-grained hierarchical information among joints. As shown in the left side of Fig.~\ref{fig1}, we group joints into six hierarchies based on the kinematic tree depth of the human body. The experiment results in the right side of Fig.~\ref{fig1} show a rising hierarchical accumulation error when the hierarchy increases from 1 to 5. \textbf{(4) Training Efficiency.}
Overly complex hierarchical modeling introduces redundant computational costs.

To solve the problems mentioned above, we first propose FastDDHPose, which consists of three key designs. \emph{First}, we disentangle the 3D pose into bone length and bone direction. This allows the diffusion model to learn their latent distributions in a lower-dimensional and structurally well-defined space, making the underlying data manifold easier to capture. \emph{Second}, we introduce the disentangled method in the forward process of diffusion model instead of decomposing the 3D HPE task into bone length and bone direction prediction task, which mitigating the amplified hierarchical reconstruction errors. \emph{Third}, For better modeling the hierarchical relation among joints, we propose KHSTDenoiser, which contains Kinematic-Hierarchical Spatial and Temporal Transformer (KHST and KHTT). KHST and KHTT make the joints pay more attention to their hierarchical-related joints, which consequently improves performance on higher-hierarchy joints and contributes to overall performance. 

% \emph{Particularly for efficiency}, compared with DDHPose, DDHPose++ simplifies the computational overhead introduced by overly complex hierarchical modeling while preserving competitive accuracy. Specifically, DDHPose++ reduces the parameter size by \textbf{9.6\%}, shortens the training time by \textbf{39.1\%}, and decreases the computational cost (GFLOPs) by \textbf{7\%}. Despite the substantially reduced model complexity, DDHPose++ achieves improved performance on Human3.6M (MPJPE \textbf{39.6 mm} vs. \textbf{39.7 mm}).

% Beyond model design, training efficiency in 3D human pose estimation is also constrained by suboptimal and non-scalable training pipelines. To address this issue, Fast3DHPE integrates Distributed Data Parallel (DDP) and Automatic Mixed Precision (AMP) to accelerate training, which leads to significant speedups across all methods.

\emph{Particularly for efficiency}, we address training efficiency at the model level. Compared with DDHPose, FastDDHPose simplifies the overly complex hierarchical modeling, thereby reducing computational overhead while preserving competitive accuracy. Specifically, FastDDHPose reduces the parameter size by \textbf{9.6\%}, shortens the training time by \textbf{39.1\%}, and decreases the computational cost (GFLOPs) by \textbf{7\%}. Despite the substantially reduced model complexity, FastDDHPose achieves improved performance on Human3.6M (MPJPE \textbf{39.6 mm} vs. \textbf{39.7 mm}).

% While recent advances have led to strong performance under controlled experimental settings, future progress in 3D human pose estimation increasingly depends on scalable, efficient, and reproducible research infrastructures. To this end, Fast3DHPE provides an efficient and user-friendly pipeline for model construction, training, and evaluation, with support for Distributed Data Parallel (DDP) multi-GPU training. This design enables efficient and scalable multi-GPU acceleration, significantly reducing training cost, and lowers the barrier to deploying large-scale experiments. By offering a flexible and efficient framework under unified experimental settings, Fast3DHPE facilitates the development, evaluation, and fair comparison of future 3D HPE methods, and we hope it will serve as a solid foundation for advancing open, reproducible, and scalable research in this field.

% Motivated by the significant progress and research potential of 3D human pose estimation, as well as its growing importance in real-world and in-the-wild applications, we aim to bridge the past and the future.

In conclusion, our contributions can be summarized as follows:
\begin{itemize}
\item We build \textbf{Fast3DHPE}, a unified and standardized framework that integrates mainstream 3D HPE methods under a consistent pipeline, providing efficient training for fairer and more reproducible benchmarking.
\item We propose \textbf{FastDDHPose}, the first disentangled diffusion-based 3D human pose estimation method that incorporates hierarchical information into both the forward diffusion and reverse denoising processes. Specifically, FastDDHPose disentangles bone length and bone direction during the forward diffusion based on the kinematic hierarchy, enabling more effective modeling of explicit pose priors. In the reverse process, we introduce the \textbf{KHSTDenoiser}, which consists of a Kinematic-Hierarchical Spatial Transformer (KHST) and a Kinematic-Hierarchical Temporal Transformer (KHTT), to strengthen joint relations by enhancing attention to kinematically adjacent joints while reducing redundant computational overhead.

\item Within the efficient Fast3DHPE framework, our improved FastDDHPose achieves state-of-the-art performance on 3D HPE benchmarks with a trade-off between accuracy and efficiency, surpassing existing disentangle-based, non-disentangle-based, and probabilistic methods by \textbf{10.2\%}, \textbf{3.4\%}, and \textbf{2.0\%}, respectively.

\item In addition, Fast3DHPE supports qualitative visualization on in-the-wild videos, which allows for intuitive inspection of model behavior in unconstrained and challenging real-world scenarios. These visual results demonstrate that FastDDHPose produces more stable and coherent pose predictions under realistic conditions. We hope this framework encourages future research to place greater emphasis on the robustness and real-world applicability of 3D human pose estimation.
\end{itemize}

This paper serves as an extension of our previous research~\cite{cai2024disentangled}. Specifically, we improve our work from three
folds: 
(1) Fast3DHPE is developed. We build a unified, standardized, and extensible framework that integrates mainstream 3D HPE approaches into a consistent pipeline. This framework provides more efficient training and fairer benchmarking. (2) FastDDHPose is proposed. We upgrade the original DDHPose by introducing KHSTDenoiser, which maintains the model’s focus on kinematic joint hierarchies while alleviating the computational overhead caused by excessively deep hierarchical modeling during the reverse diffusion process in DDHPose~\cite{cai2024disentangled}. (3) We provide more empirical results to complement quantitative evaluations, together with extensive qualitative visualizations on both in-the-lab and in-the-wild scenarios. These visual results offer intuitive insights into model behavior and demonstrate the robustness and stability of our method under both controlled benchmarks and challenging real-world conditions.

\section{Related Work}\label{sec2}
\subsection{3D Human Pose Estimation}
3D HPE can be divided into two categories, one that directly regresses the 3D human pose from raw RGB images~\cite{zhou2016deep,pavlakos2017coarse,pavlakos2018ordinal} and another that first detects the 2D human pose from RGB images by using one of the 2D human pose estimation methods like HRNet~\cite{wang2020deep}, CPN~\cite{chen2018cascaded}, OpenPose~\cite{cao2017realtime}, AlphaPose~\cite{fang2017rmpe} and then make a 2D-to-3D lifting to get the final estimation results. ~\cite{tekin2016direct,pavlakos2017coarse,sun2018integral} directly use convolutional neural network to regress 3D pose from a feature volume. 

Based on the accuracy improvement of 2D human pose estimation, VideoPose3D~\cite{pavllo:videopose3d:2019} uses a fully convolutional model based on dilated temporal convolutions to estimate 3D poses and achieves better results. In addition, several works~\cite{li2023pose,li2025graphmlp,zhao2019semantic,pavlakos2018ordinal,yu2023gla,xu2021graph,mehraban2024motionagformer,peng2024dual,cai2025nanohtnet} model 3D human pose estimation from a graph-based perspective, where joints are represented as graph nodes and skeletal connections are encoded as edges. These methods leverage the inherent skeletal structure to capture joint dependencies. Besides, ~\cite{zheng20213d, Zhang_2022_CVPR, Zhao_2023_CVPR,li2025h,li2024hourglass} demonstrate that 3D poses in the video can be effectively estimated with spatial-temporal transformer architecture. Due to the superior performance of two-stage methods, we also employ a two-stage approach for 3D human pose estimation in this paper. While these models are capable of exploring spatial-temporal context information, they always fail to incorporate fine-grained hierarchical information. This leads to a higher hierarchical accumulation error from hierarchy 1 to hierarchy 5 in the right portion of Fig.\ref{fig1}. Therefore, we apply KHST and KHTT in our method, providing more hierarchical features for better modeling.

\subsection{Diffusion Model}
The diffusion model belongs to a class of generative models~\cite{ng2001discriminative,kingma2013auto,lecun2006tutorial,goodfellow2014generative}, which has outstanding performance in image and video generation~\cite{batzolis2021conditional, nichol2021glide, ho2022cascaded,ji2025layerflow}, image super-resolution~\cite{saharia2022image}, semantic segmentation~\cite{baranchuk2021label}, multi-modal tasks~\cite{fan2023frido} and so on.  The diffusion model is first introduced by ~\cite{sohl2015deep}, which defines two stages which are the forward process and the reverse process. The forward process refers to the gradual addition of Gaussian noise to the data until it becomes random noise, while the reverse process is the denoising of noisy data to obtain the true samples. The following works DDPM~\cite{ho2020denoising} and DDIM~\cite{song2020denoising} simplify and accelerate previous diffusion models which make a solid foundation in this area. 

Recent explorations~\cite{choi2022diffupose,holmquist2022diffpose,ci2023gfpose,shan2023diffusion} try to apply the diffusion model to 3D human pose estimation. Note that~\cite{gong2023diffpose} also uses a diffusion model for 3D HPE, but they additionally introduce the heatmap distribution of 2D pose, and the depth distribution to initialize 3D pose distribution, making a GMM-based forward diffusion process, so that they have a better performance than the other diffusion-based 3D HPE model. However, these approaches directly add $t$-step noise in the forward process to the original 3D pose, which is not conducive to learning the explicit human pose prior. Additionally, some methods~\cite{xu2020deep,chen2021anatomy,wang2022motion} have a higher accumulation of errors that disentangle the 3D joint location to the prediction of bone length and bone direction. We introduce the disentanglement strategy in the forward process of the diffusion model, integrating the explicit human body prior to the diffusion model, and proposing the first disentangle-based diffusion model for 3D HPE. As a result, we achieve outstanding results on 3D HPE benchmarks.

\section{Fast3DHPE}\label{sec4}
\begin{figure*}[t]
    \centering
    \includegraphics[width=\textwidth]{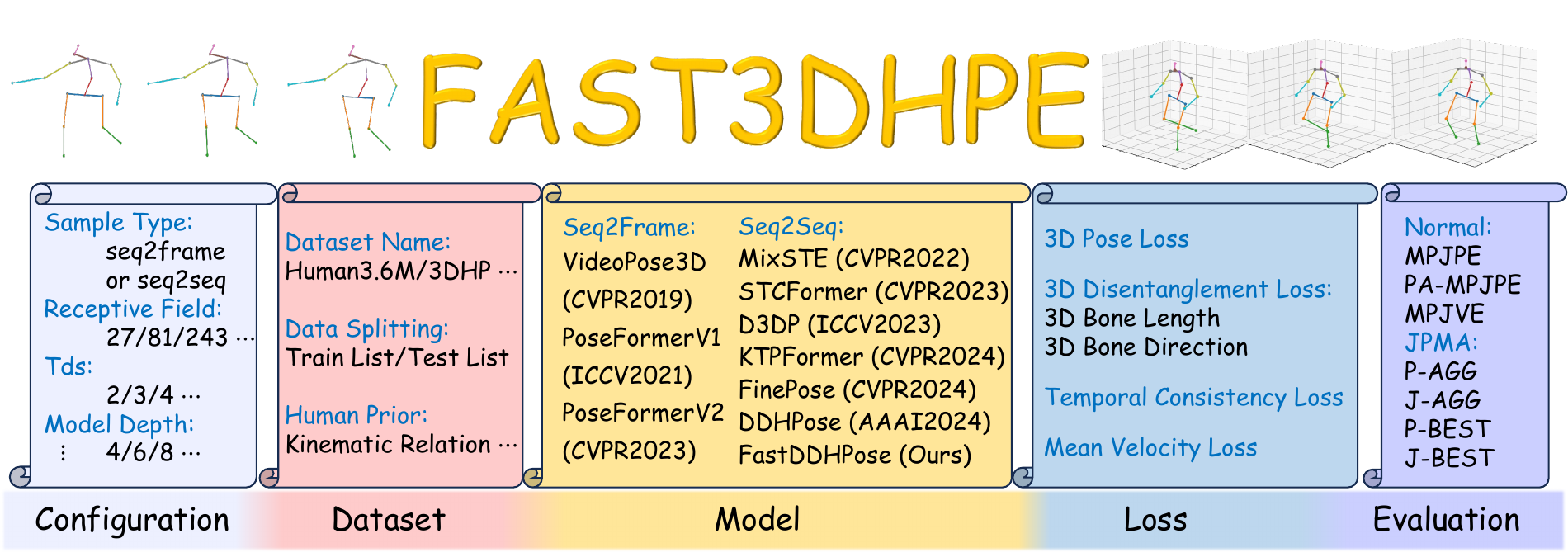} % Reduce the figure size so that it is slightly narrower than the column. Don't use precise values for figure width.This setup will avoid overfull boxes.
    \caption{The overview of Fast3DHPE, which consists of five functional modules.}
    \label{fig2}
\end{figure*}
\subsection{Overview}
To enable fair comparison across different methods and improve training efficiency in 3D human pose estimation, we introduce Fast3DHPE, an open-source and extensible toolbox for 3D human pose estimation built upon the PyTorch~\cite{paszke2019pytorch} deep learning framework. The toolbox provides a unified implementation of a broad spectrum of state-of-the-art algorithms for 3D pose estimation and supports widely-used benchmarks. In contrast to existing projects that typically focus on reproducing a single method, Fast3DHPE integrates multiple representative and competitive approaches into a coherent framework. This design not only encapsulates recent methodological advances but also incorporates common best practices, thereby enabling fair, systematic, and reproducible comparisons in terms of both effectiveness and computational efficiency.

Fast3DHPE includes numerous pretrained models, standardized training and evaluation pipelines, and comprehensive benchmark results. These provide valuable empirical insights and serve as reliable references for future investigations. Owing to its modular architecture and diverse algorithmic choices, Fast3DHPE offers researchers a flexible platform to rapidly prototype new ideas, conduct empirical studies, and advance methodological development in the field.

Beyond these general capabilities, Fast3DHPE is designed around several key principles that significantly enhance usability, reproducibility, and research efficiency:

\begin{enumerate}
    \item \textbf{Unified settings.} Fast3DHPE organizes diverse preprocessing pipelines and evaluation protocols within a normalized and configurable framework, allowing researchers to investigate the effects of model designs or other parameters under controlled experimental conditions.

    \item \textbf{Modular design.} The framework is structured into clearly defined functional modules, enabling researchers to flexibly assemble new components or introduce novel ideas without modifying the entire system.

    \item \textbf{Rich model zoo.} Fast3DHPE provides a rich and representative collection of 3D human pose estimation models, together with pretrained weights and detailed performance reports on mainstream 3D HPE benchmarks. These resources enable systematic evaluation and fair comparison across methods, offering strong empirical references for subsequent research.
    \footnotetext[1]{PyTorch DDP tutorial: \url{https://pytorch.org/tutorials/intermediate/ddp_tutorial.html}}
    \footnotetext[2]{PyTorch AMP tutorial: \url{https://pytorch.org/tutorials/recipes/recipes/ amp_recipe.html}}
    \item \textbf{Efficient execution.} The implementation leverages multi-GPU Distributed Data Parallel (DDP)\footnotemark{} training and Auto Mixed Precision (AMP)\footnotemark{}, enabling faster training, reduced memory usage, thereby facilitating more efficient model iteration in practical applications as well as academic experiments and large-scale deployments.
\end{enumerate}

\subsection{Architecture}
The overall architecture of Fast3DHPE is shown in Fig.~\ref{fig2}, which is organized into five functional modules designed to provide a unified, extensible, and reproducible framework for 3D human pose estimation.

% \textbf{(2) Dataset} standardizes data preprocessing, augmentation, and dataset splitting, while also supplying human-body prior information such as skeleton definitions, joint mappings, and symmetry annotations.
% \textbf{(3) Model} integrates a wide collection of representative and state-of-the-art architectures~\cite{pavllo:videopose3d:2019,zheng20213d,Zhao_2023_CVPR,li2022mhformer,tang20233d,shan2023diffusion,peng2024ktpformer,xu2024finepose,cai2024disentangled}, offering a flexible interface for instantiating backbone networks and assembling customized pipelines. 
% \textbf{(4) Loss} provides a suite of widely used objective functions for supervised training, enabling consistent comparison across different methods and training regimes. 
% \textbf{(5) Evaluation} supports multiple evaluation protocols and metrics commonly used in 3D pose estimation, facilitating standardized and comprehensive performance assessment across datasets and algorithms. To further enhance the interpretability of model predictions, Fast3DHPE includes a complete set of visualization utilities that render the estimated 3D poses and support qualitative comparisons among different algorithms.

\subsubsection{Configuration}
Configuration manages all experiment-level settings, including model specifications, dataset definitions, optimization strategies, and training schedules, ensuring transparent and reproducible experimental control. 
For clarity, we summarize a few of the most important configurations here.

\begin{enumerate}
    \item \textbf{Sample type.}  
One of the most critical configuration settings in Fast3DHPE is the \emph{sample type}, which determines how temporal 2D pose sequences are mapped to 3D pose predictions. Fast3DHPE supports two commonly used paradigms. 
\textbf{seq2frame.} 
Early 3D pose estimation approaches~\cite{pavllo:videopose3d:2019,Zhao_2023_CVPR,zheng20213d,li2022mhformer} typically adopt a seq2frame strategy, where a temporal window of 2D poses (e.g., 27/81/243 frames) is used to regress the 3D pose of a single target frame. Although effective, this design requires running the model repeatedly with heavily overlapping windows in order to obtain the full 3D sequence. As a result, seq2frame introduces substantial redundant computation and leads to high overall inference cost.
\textbf{seq2seq.} 
Recent works~\cite{Zhang_2022_CVPR,shan2023diffusion,tang20233d,peng2024ktpformer,xu2024finepose,cai2024disentangled,gong2023diffpose,choi2022diffupose} increasingly favor seq2seq formulations, in which an input window of 2D poses is mapped directly to a 3D pose sequence of the same length. By predicting all frames within the window simultaneously, seq2seq eliminates redundant forward passes across overlapping windows and significantly reduces the total inference cost, resulting in much higher computational efficiency.
\item \textbf{Receptive Field.}
Estimating 3D human pose solely from 2D joint sequences is inherently an ill-posed problem, as depth ambiguities cannot be resolved from a single frame. Current approaches overcome this limitation by exploiting temporal cues, where consistency across adjacent frames and human-body priors help compensate for the missing depth information. The \emph{receptive field} specifies the temporal window size through which the model captures such dependencies, effectively determining how much contextual motion information is available for inferring 3D structure. Common receptive field settings include 27, 81, or 243 frames, each providing different levels of temporal context and influencing the model’s capacity to reason about dynamic human motion.
\item \textbf{Tds.} 
The \emph{Temporal Downsampling Strategy} (Tds) enlarges the temporal receptive field by uniformly sampling input frames at a fixed interval, allowing the model to access longer-range motion information without increasing the number of input frames or computational cost. This design was use in~\cite{shan2022p,tang20233d} and has since been widely adopted.

\end{enumerate}

\subsubsection{Dataset}
In our benchmark, we integrate the two most widely used datasets in the 3D human pose estimation community. 
For each dataset, we follow the conventional train/test splits adopted by previous work to ensure fair comparison. 
We also provide the corresponding human skeletal definitions and joint-connectivity information, enabling different categories of methods to correctly utilize the skeleton structure. 
The two datasets are introduced as follows:
\begin{enumerate}
    \item \textbf{Human3.6M}
    ~\cite{h36m_pami} is widely used in 3D HPE task. It contains 3.6 million 3D human poses and corresponding images with 11 professional actors and collected in 17 scenarios. Following the previous work~\cite{pavllo:videopose3d:2019,zheng20213d,Zhang_2022_CVPR}, we use S1-S9 for training and use S9 and S11 for testing.

    \item \textbf{MPI-INF-3DHP}
    ~\cite{mehta2017monocular} record 8 actors, composed of 4 males and 4 females, each undertaking 8 different sets of activities. We use eight activities performed by eight actors to train our model, while the test dataset has seven different activities.

\end{enumerate}

\subsubsection{Model}
In this work, we integrate ten representative 3D human pose estimation models, 
including nine mainstream approaches and our proposed FastDDHPose. 
These models cover a broad spectrum of design philosophies and can be categorized along two orthogonal dimensions.

\begin{enumerate}
    \item \textbf{Sampling strategy.}
    According to how the models process temporal information, they can be divided into two groups:
    \emph{seq2frame} methods~\cite{pavllo:videopose3d:2019,zheng20213d,Zhao_2023_CVPR}, which estimate the 3D pose of the center frame from a temporal window of 2D inputs, 
    and \emph{seq2seq} methods~\cite{Zhang_2022_CVPR,tang20233d,shan2023diffusion,peng2024ktpformer,xu2024finepose,cai2024disentangled}, which predict a sequence of 3D poses from an entire 2D pose sequence.

    \item \textbf{Output numbers.}
    Based on whether the model predicts a single deterministic pose or multiple plausible hypotheses, the integrated methods include:
    \emph{deterministic} models~\cite{pavllo:videopose3d:2019,zheng20213d,Zhao_2023_CVPR,Zhang_2022_CVPR,tang20233d} that output a single 3D pose estimate, 
    and \emph{probabilistic} models~\cite{shan2023diffusion,peng2024ktpformer,xu2024finepose,cai2024disentangled} that generate multiple hypotheses to capture pose ambiguity.

\end{enumerate}

\subsubsection{Loss Function}
The loss function consists of four components: a standard 3D pose regression loss, a disentanglement loss that explicitly supervises bone lengths and directions, a temporal consistency loss, and a mean velocity loss that enforces motion-level consistency across frames.
\begin{enumerate}
    \item \textbf{3D Pose Loss.}
    To directly constrain the denoised 3D pose predicted by our model, we employ the standard and most widely used 3D pose regression loss $\ell_{pos}$ in 3D human pose estimation, formulated as the $\ell_2$ distance between the estimated pose $\tilde{y}_{0}$ and the ground-truth pose $y_{0}$:
    \begin{equation}
        \ell_{pos} = \left \| \tilde{y}_{0} - y_{0} \right \|_{2}
    \end{equation}

    \item \textbf{3D Disentanglement Loss.}
    3D disentanglement loss is utilized in~\cite{chen2021anatomy,cai2024disentangled} to aid the model in learning the explicit priors during the forward diffusion process. Given the 3D ground truth pose sequence $y_{0}$ and the predicted 3D pose sequence $\tilde{y}_{0}$, we decompose $y_{0}$ to bone length $l_{0}$ and bone direction $d_{0}$. Similarly, we can obtain the disentangled bone length prediction $\tilde{l}_{0}$ and bone direction prediction $\tilde{d}_{0}$. And for the $i$-th bone, length $l_{0}^{i}$, $\tilde{l}_{0}^{i}$ and direction $d_{0}^{i}$, $\tilde{d}_{0}^{i}$ are defined as:

    \begin{equation}
    \begin{aligned}
    l_{0}^{i} &= \left \| y_{0}^{c_{i}} - y_{0}^{p_{i}} \right \|_{2}, &\tilde{l}_{0}^{i} &= \left \| \tilde{y}_{0}^{c_{i}} - \tilde{y}_{0}^{p_{i}} \right \|_{2}\\
    d_{0}^{i} &= \frac{y_{0}^{c_{i}} - y_{0}^{p_{i}}}{\left \| y_{0}^{c_{i}} - y_{0}^{p_{i}} \right \|_{2}}, &\tilde{d}_{0}^{i} &= \frac{\tilde{y}_{0}^{c_{i}} - \tilde{y}_{0}^{p_{i}}}{\left \| \tilde{y}_{0}^{c_{i}} - \tilde{y}_{0}^{p_{i}} \right \|_{2}}
    \end{aligned}
    \end{equation}
    where $c_{i}$ and $p_{i}$ is the child joint and parent joint of the $i$-th bone. Then the disentanglement loss $\ell_{dis}$ we use in our training stage can be defined as:\\
    \begin{equation}
    \begin{gathered}
    \begin{aligned}
        \ell_{l} &= \left \| \tilde{l}_{0} - l_{0} \right \|_{2}, & \ell_{d} &= \left \| \tilde{d}_{0} - d_{0} \right \|_{2}\\
    \end{aligned} \\
    \ell_{dis} = \ell_{l} + \ell_{d}
    \end{gathered}
    \end{equation}

    \item \textbf{Temporal Consistency Loss.}
    Temporal Consistency Loss is introduced in ~\cite{hossain2018exploiting} that regularizes both frame-to-frame smoothness and velocity alignment.
    Given a predicted sequence $\tilde{y} = \{\tilde{y}_{t}\}_{t=1}^{T}$, we compute the first-order difference between consecutive predicted frames:
    \begin{equation}
    \Delta \tilde{y}_{t} = \tilde{y}_{t+1} - \tilde{y}_{t}
    \end{equation}
    To account for different perceptual importance across joints, a dataset-specific weight vector 
    $w \in \mathbb{R}^{J}$ is applied, yielding the weighted temporal smoothness loss:
    \begin{equation}
        \ell_{\text{temp}} = \mathbb{E}_{t} \left[ \, w \cdot \| \Delta \tilde{y}_{t} \|^{2} \, \right]
    \end{equation}
    
    \item \textbf{Mean Velocity Loss.}
    Mean Velocity Loss is introduced in~\cite{pavllo:videopose3d:2019}
    to explicitly enforce temporal consistency by penalizing the deviation between predicted and ground-truth joint velocities. Given a predicted sequence $\tilde{y} = \{\tilde{y}_{t}\}_{t=1}^{T}$ and the corresponding ground truth $y = \{y_{t}\}_{t=1}^{T}$, the velocity loss can be define as:
    
    \begin{equation}
        \ell_{\text{vel}} = 
    \mathbb{E}_{t} \left[ 
    \left\| (\tilde{y}_{t+1} - \tilde{y}_{t}) - (y_{t+1} - y_{t}) \right\|
    \right]
    \end{equation}

\end{enumerate}

\begin{figure*}[t]
    \centering
    \includegraphics[width=\textwidth]{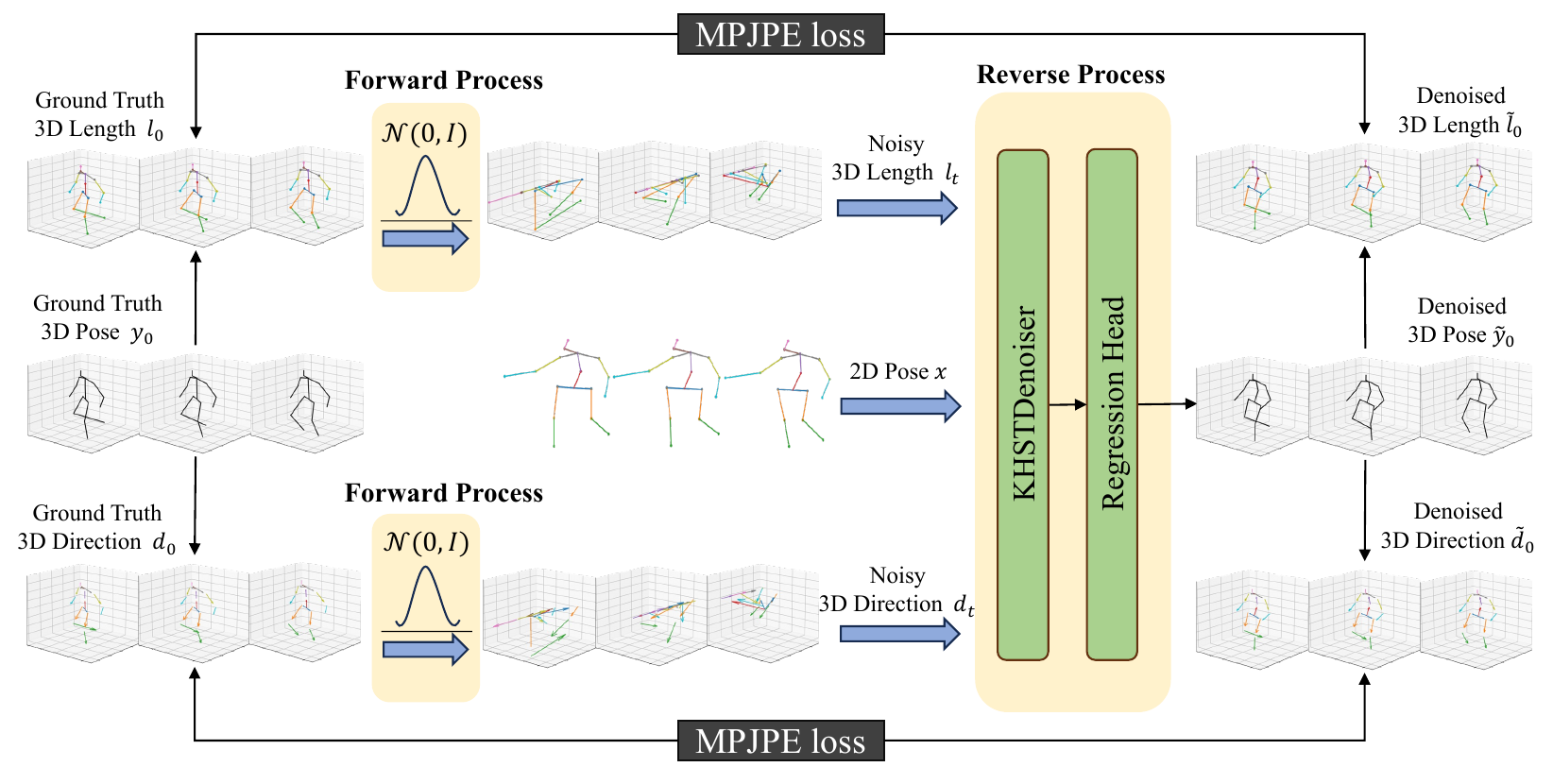} % Reduce the figure size so that it is slightly narrower than the column. Don't use precise values for figure width.This setup will avoid overfull boxes.
    \caption{The overview of FastDDHPose's training pipeline. The input consists of 2D pose, 3D bone length and 3D bone direction. For better clarity, only three frames of input are illustrated here as an example. }
    \label{fig3}
\end{figure*}

\subsubsection{Evaluation}
We evaluate our method following two commonly used evaluation protocols in 3D human pose estimation, 
corresponding to deterministic methods and probabilistic methods respectively.

\begin{enumerate}
    \item \textbf{Deterministic Evaluation Metrics.}
    For deterministic 3D pose estimators, we adopt the standard and widely used metrics in~\cite{martinez2017simple,pavllo:videopose3d:2019}:
\begin{itemize}
    \item \textbf{MPJPE} (Mean Per Joint Position Error): the mean Euclidean distance between predicted and ground-truth joint positions.
    \item \textbf{P-MPJPE} (Procrustes-Aligned MPJPE): MPJPE computed after rigid alignment using Procrustes analysis.
    \item \textbf{N-MPJPE} (Normalized MPJPE): MPJPE after applying scale normalization to the predicted pose.
    \item \textbf{MPJVE} (Mean Per Joint Velocity Error): the mean difference between predicted and ground-truth joint velocities, 
    reflecting temporal smoothness and motion consistency.
\end{itemize}

\item \textbf{Probabilistic Evaluation Metrics.}
For probabilistic 3D pose estimation methods that output a distribution or multiple samples of 3D poses, we follow the metrics that used in~\cite{shan2023diffusion}:
\begin{itemize}
    \item \textbf{P-Best}: Select the predicted 3D pose hypothesis that is closest to the ground truth.
    \item \textbf{J-Best}: For each joint, select the hypothesis closest to the ground truth and compose them into a final pose.
    \item \textbf{P-Agg}: Aggregate predictions at the pose level, treating each 3D pose as the smallest unit to form the final output.
    \item \textbf{J-Agg}: Aggregate predictions at the joint level, treating each joint as the smallest unit to form the final pose and capture finer distributional consistency.
\end{itemize}
\end{enumerate}

\section{FastDDHPose}\label{sec3}
Within Fast3DHPE, we introduce FastDDHPose, which leverages the strong latent distribution modeling capability of diffusion models to explicitly model the distributions of bone length and bone direction while avoiding further amplification of hierarchical error accumulation.

The overview of our proposed \textbf{FastDDHPose} is in Fig.~\ref{fig3}. In our framework, we decompose the 3D joint location into the bone length and bone direction, adding noise in the forward process. After the forward process, the noisy bone length, noisy bone direction, and 2D pose are fed to \textbf{KHSTDenoiser}, which contains Kinematic-Hierarchical Spatial and Temporal Transformer to reverse the 3D pose from the noisy input. Further details will be introduced in the following section.

\subsection{Disentanglement Strategy}
We first introduce the motivation of why we use the disentanglement strategy in our paper. Modeling the distribution of full 3D poses directly in diffusion-based frameworks is intrinsically challenging. A human pose lies in a high-dimensional and kinematically articulated space, where different motion patterns (e.g., walking, phoning, sitting) exhibit diverse joint dependencies. Learning such complex and dense correlations among all joints often makes the diffusion process harder to optimize and less stable. Although several previous works attempt to introduce explicit kinematic priors through disentanglement, they typically require the model to predict multiple components. This multi-branch prediction tends to accumulate hierarchical errors and may compromise final pose accuracy.

% The original non-disentangle diffusion-based methods directly take the 3D joint sequence as input without any skeleton structural prior. Modeling the dependencies among each joint pair tends to be challenging due to their complex and dense relation which makes the optimization task more difficult. 

Motivated by the above analysis, we first alleviate the difficulty of modeling full 3D poses by decomposing the ground-truth pose $y_{0}\in\mathbb{R}^{N\times J\times 3}$ into two low-dimensional components: bone length $l_{0}\in\mathbb{R}^{N\times (J-1)\times 1}$ and bone direction $d_{0}\in\mathbb{R}^{N\times (J-1)\times 3}$. For the $i$-th bone, ground truth length $l_{0}^{i}$ and direction $d_{0}^{i}$ can be defined as:
\begin{equation}
    l_{0}^{i} = \left \| y_{0}^{c_{i}} - y_{0}^{p_{i}} \right \|_{2},\quad d_{0}^{i} = \frac{y_{0}^{c_{i}} - y_{0}^{p_{i}}}{\left \| y_{0}^{c_{i}} - y_{0}^{p_{i}} \right \|_{2}}
    \end{equation}
Here, $N$ denotes the sequence length and $J$ the number of joints, $c_{i}$ and $p_{i}$ are the child joint and parent joint, which are in the upstream and downstream of the $i$-th bone according to the forward kinematic structure defined in the left portion of Fig.~\ref{fig1}. This decomposition reformulates the original dense joint-position modeling problem into two sparse and substantially lower-dimensional subspaces, which significantly reduces the modeling complexity and stabilizes diffusion optimization. In particular, the disentangled bone-length component exhibits strong identity consistency, further providing a stable structural cue for the diffusion process. 

While the above decomposition already reduces the complexity of the diffusion space, an equally important consideration is to avoid predicting too many disentangled variables during regression, which would otherwise magnify hierarchical cumulative errors. To this end, the disentangled bone length and bone direction are modeled separately during the forward diffusion process, while they are jointly integrated in the reverse process to produce the final prediction.

% But in our approach, Our disentangle-based method first decomposes ground truth 3D pose $y_{0} \in\ \mathbb{R}^{N\times J\times3}$ to bone length $l_{0} \in\ \mathbb{R}^{N\times (J-1)\times1}$ and bone direction $d_{0} \in\ \mathbb{R}^{N\times (J-1)\times3}$, where $N$ is the frame length of the input sequences, $J$ is the number of joints. This operation divides the dense and high-dimensional problem into multiple sparse and low-dimensional sub-problems, making the gradient-based optimization easier. 

% Besides, The disentangling representation with bone length and direction makes it easier to add structural constraints, such as temporal consistency in bone length. The addition of bone length loss as a constraint enhances output certainty and shows effectiveness in the experiment.

\subsection{The Forward Process}
The forward process is an approximate posterior that follows a Markov chain gradually adding Gaussian noise $\mathcal{N}(0,I)$ to the original data $x_{0}$. 
Following DDPM~\cite{ho2020denoising}, the forward process is defined as:
\begin{equation}
q(x_{t} \mid x_{0}) = \mathcal{N}\!\left(x_{t}; \sqrt{\bar{\alpha}_{t}}\, x_{0}, (1-\bar{\alpha}_{t})I \right)
\label{eq:forward}
\end{equation}
where $\bar{\alpha}_{t}= \prod_{s=0}^{t} \alpha_{s}$ and $\alpha_{s}=1-\beta_{s}$. 
Here, $\beta_{s}$ is the noise schedule, and we adopt the cosine schedule in~\cite{song2020improved}, which monotonically increases with the timestep $t$.

\par During the training stage in Fig.~\ref{fig3}, when we get the disentangled bone length $l_{0}$ and bone direction $d_{0}$, we can do the forward process separately in Eq (2) to get the noisy bone length $l_{t}$ and bone direction $d_{t}$ by adding $t$-step Gaussian noise as:
\begin{equation}
    \begin{aligned}
    l_{t} &= \sqrt{\bar{\alpha}_{t}}\, l_{0} + \sqrt{1-\bar{\alpha}_{t}}\, \epsilon \\
    d_{t} &= \sqrt{\bar{\alpha}_{t}}\, d_{0} + \sqrt{1-\bar{\alpha}_{t}}\, \epsilon
    \end{aligned}
    \end{equation}
where $\epsilon$ is the random Gaussian sampled at the $t$-step. 

\subsection{The Reverse Process} 
% \subsubsection{Overview}
In the training stage shown in Fig.~\ref{fig3}, given a 2D pose sequence 
\( x \in \mathbb{R}^{N \times J \times 2} \),
the contaminated bone length \( l_t \) and bone direction \( d_t \)
from the forward process are concatenated as 

\begin{equation}
    z_t = \operatorname{Concat}(x,\ l_t,\ d_t)
    \label{eq:concat_input}
    \end{equation}

The fused representation is fed into the Kinematic-Hierarchical Spatial and Temporal Denoiser
(\textbf{KHSTDenoiser}) $\mathcal{D}_\theta$ followed by a regression head $g_\theta$,
yielding the estimated clean 3D pose $\tilde{y}_0 = g_\theta\!\left( \mathcal{D}_\theta(z_t) \right)$.
A detailed description of KHSTDenoiser is provided in Sec.~\ref{sec332}.

\begin{figure}[t]
    \centering
    \includegraphics[width=1\columnwidth]{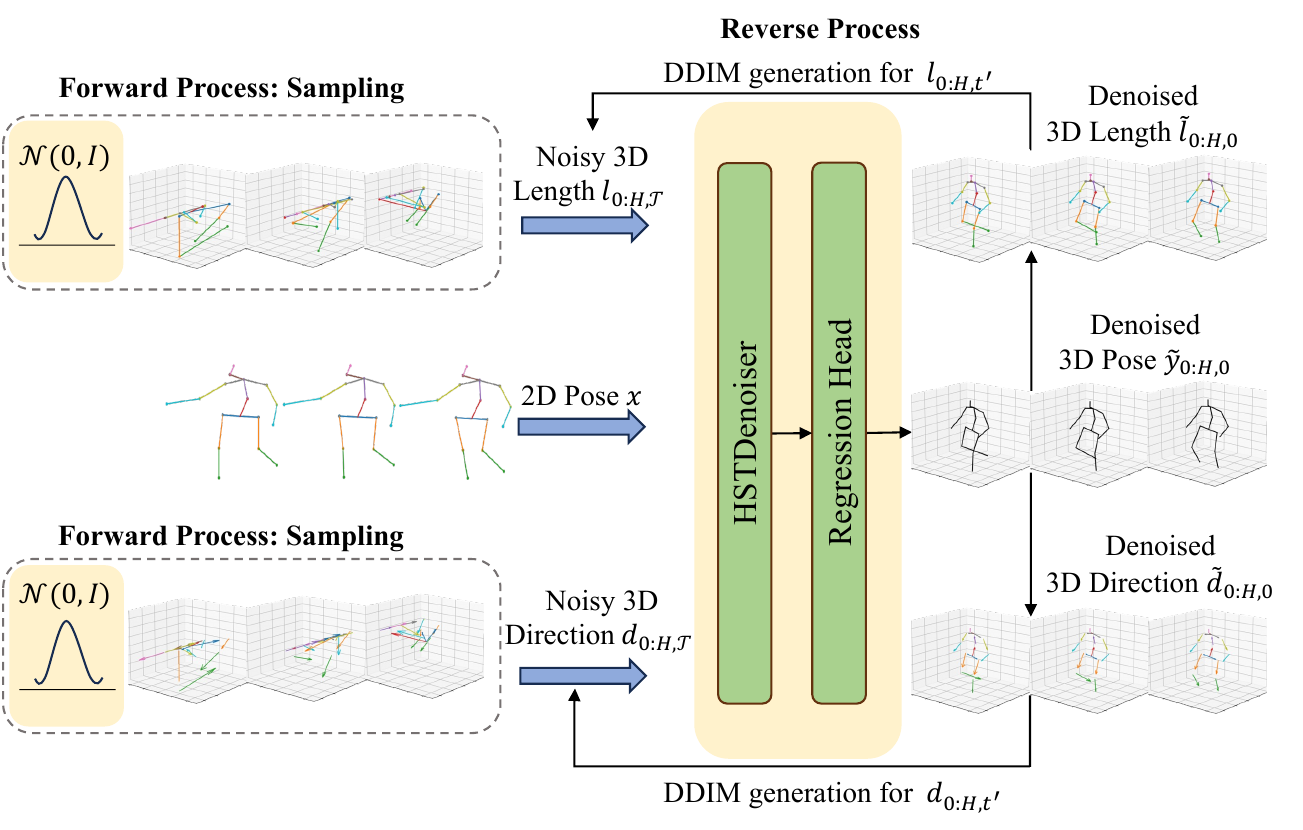} % Reduce the figure size so that it is slightly narrower than the column. Don't use precise values for figure width.This setup will avoid overfull boxes.
    \caption{The overview of the inference pipeline. The input consists of 2D pose, 3D bone noisy length and 3D bone noisy direction. For better clarity, only three frames of input are illustrated here as an example.}
    % \vspace{-3.85ex}
    \label{fig4}
    \end{figure}

At the inference stage shown in Fig.~\ref{fig4}, inspired by D3DP~\cite{shan2023diffusion}, we first simultaneously sample $H$ hypotheses from a Gaussian distribution as the initial noisy bone length and direction. For each hypothesis, the noisy pair at step $t$ is fed, together with the 2D pose sequence, into the trained KHSTDenoiser and the regression head to obtain an estimated 3D pose $\tilde{y}_{0:H,0}$. We then disentangle $\tilde{y}_{0:H,0}$ into bone length and bone direction, denoted as $\tilde{l}_{0:H,0}$ and $\tilde{d}_{0:H,0}$, which serve as the clean estimates at step $t=0$ and act as the initial state for the subsequent DDIM~\cite{song2020denoising} iteration. 
Based on these clean estimates at step $t$, DDIM is then applied to produce the corresponding noisy samples $\tilde{l}_{0:H,t'}$ and $\tilde{d}_{0:H,t'}$ for the next step $t'$:

\begin{equation}
    \begin{aligned}
    \tilde{l}_{0:H,t'} &= c_{t'}\,\tilde{l}_{0:H,0}
        + s_{t'}\,\epsilon_{tl}
        + \sigma_t \epsilon \\
    \tilde{d}_{0:H,t'} &= c_{t'}\,\tilde{d}_{0:H,0}
        + s_{t'}\,\epsilon_{td}
        + \sigma_t \epsilon
    \end{aligned}
    \end{equation}
where $c_{t'}=\sqrt{\bar\alpha_{t'}}$ and $s_{t'}=\sqrt{1-\bar\alpha_{t'}-\sigma_t^2}$, $\epsilon_{t l}$ and $\epsilon_{t d}$ denote the normalized noise at step $t$,
\begin{equation}
\begin{aligned}
\epsilon_{t l} &= 
\frac{\tilde{l}_{0:H,t}-\sqrt{\bar{\alpha}_{t}}\,\tilde{l}_{0:H,0}}
{\sqrt{1-\bar{\alpha}_{t}}} \\
\epsilon_{t d} &= 
\frac{\tilde{d}_{0:H,t}-\sqrt{\bar{\alpha}_{t}}\,\tilde{d}_{0:H,0}}
{\sqrt{1-\bar{\alpha}_{t}}}
\end{aligned}
\end{equation}

We repeat this procedure for $W$ denoising steps, obtaining $H$ candidate 3D poses, and finally apply the JPMA~\cite{shan2023diffusion} method to select the optimal prediction. Appropriately increasing the hypothesis number $H$ and the iteration times $W$ yields more accurate bone length and direction estimates, which further improves MPJPE and P-MPJPE in our experiments.
\begin{figure*}[t]
    \centering
    \includegraphics[width=\textwidth]{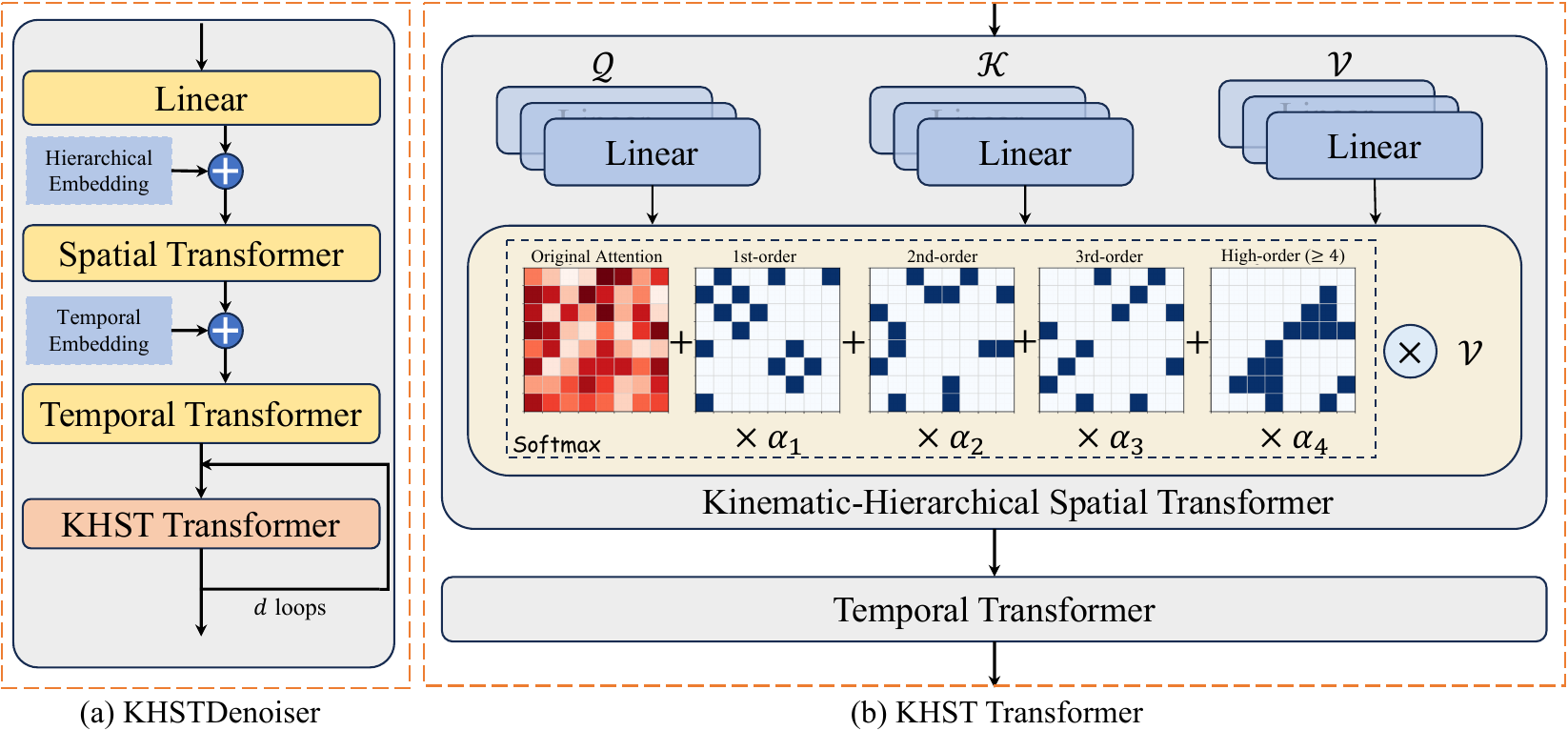} % Reduce the figure size so that it is slightly narrower than the column. Don't use precise values for figure width.This setup will avoid overfull boxes.
    \caption{(a): The overview of KHSTDenoiser. Hierarchical embedding and Temporal Embedding are used in the spatial-temporal transformer to better model the hierarchical relation of spatial position information and temporal information. (b): The architecture of our KHSTDenoiser, which contains Kinematic-Hierarchical Spatial and Temporal Transformer. }
    \label{fig5}
\end{figure*}

% (b): The architecture of our KHSTDenoiser, which contains Kinematic-Hierarchical Spatial and Temporal Transformer. $f$ is the feature extracted by the spatial-temporal transformer and $f_{c}$ is the child joint feature separated from $f$. The input consists of $N$ frames for both 2D pose and 3D pose.

\subsubsection{Kinematic-Hierarchical Spatial and Temporal Denoiser}\label{sec332}
% \paragraph{Kinematic-Hierarchical Spatial and Temporal Denoiser}
Both in the training or inference phase, noisy bone length and bone direction are fed into our KHSTDenoiser to reconstruct the original data. KHSTDenoiser consists of Kinematic-Hierarchical Spatial and Temporal Transformer (KHST and KHTT), which is used to explore the hierarchical information, specifically the relation among the joint, the parent joint, and the child joint. The main architecture is shown in Fig.~\ref{fig5}. 

We utilize a linear layer to enhance the input feature and use the spatial-temporal transformer block in MixSTE~\cite{Zhang_2022_CVPR} to extract joint features. We also introduce Hierarchical spatial position Embedding (HiE) for better spatial position modeling and Temporal Embedding for better temporal relation modeling. In particular, HiE not only contains the spatial position information of each joint but also contains the joint hierarchy information. We split the joints into six hierarchies according to the joint's depth of the human body tree-like structure to build hierarchical embedding, which is shown in the left portion of Fig.~\ref{fig1}. It means the joints in the same hierarchy share the same embedding. Based on hierarchical embedding, the hierarchical-related information can be well learned by our model. After one layer of spatio-temporal transformer modeling, we utilize the KHST and KHTT, which we introduce in the subsequent section, to model the spatio-temporal correlations of joints through $d$ loops alternately.

\paragraph{KHST}
In the KHST module, we enhance spatial relational modeling by injecting adaptive kinematic-hierarchical priors into the attention mechanism. 
Given the input features, we first project them into query $Q$, key $K$, and value $V$ using linear layers, and compute the original attention map through the scaled dot-product~\cite{Vaswani_Shazeer_Parmar_Uszkoreit_Jones_Gomez_Kaiser_Polosukhin_2017} and softmax:
\begin{equation}
    A_{\text{orig}} = \text{Softmax}\left(\frac{QK^{\top}}{\sqrt{d}}\right) \in \mathbb{R}^{J \times J}
    \label{eq:orig_attention}
    \end{equation}

To incorporate human-body kinematic structure, we predefine several hierarchical adjacency matrices that encode different orders of joint relations, including: (1) first-order parent--child neighbors, 
(2) second-order neighbors, (3) third-order neighbors, and (4) high-order ($\ge 4$) long-range dependencies.
Let $\{A_1, A_2, A_3, A_4\}$ denote these hierarchical attention bases. 
KHST learns a set of adaptive mixing coefficients $\{\alpha_1, \alpha_2, \alpha_3, \alpha_4\}$ to aggregate these priors. 
The refined spatial attention matrix is formulated as:
\begin{equation}
    A = A_{\text{orig}}
        + \alpha_1 A_1
        + \alpha_2 A_2
        + \alpha_3 A_3
        + \alpha_4 A_{\ge 4}
    \label{eq:hier_attention}
    \end{equation}

This adaptive hierarchical integration enables the model to dynamically emphasize appropriate kinematic dependencies based on the input pose representation and diffusion timestep, leading to a more structured and interpretable spatial attention pattern. The refined attention matrix is then multiplied with $V$ to obtain the spatial-enhanced joint representation, which is fed into the subsequent temporal transformer.

\paragraph{KHTT}
Following the spatial enhancement in KHST, the Kinematic-Hierarchical Temporal Transformer (KHTT) is introduced to further strengthen the temporal correlations among kinematic related joints. 

As a simple yet effective design, KHTT adopts a standard temporal transformer along the frame dimension, producing a \emph{frame-to-frame} attention map $A_{\text{temp}} \in \mathbb{R}^{N \times N}$, which captures the motion dynamics of each joint across time and propagates temporal information among frames. 
This temporal modeling plays a critical role in preserving and propagating the strengthened hierarchical dependencies obtained in the spatial stage.

\paragraph{Summary}
By sequentially combining KHST and KHTT, the model not only learns structured spatial relations from the human kinematic hierarchy but also reinforces these relations over time, enabling more consistent interaction among joints that share strong kinematic dependencies.

% \paragraph{Transformer}
% The Transformer model we used in our approach is followed by ~\cite{Vaswani_Shazeer_Parmar_Uszkoreit_Jones_Gomez_Kaiser_Polosukhin_2017}, the basic idea of query, key, and value is that the query is used to match with the key, and then according to the degree of matching, to selectively focus on the value. This design allows the output to selectively pay attention to the value based on the query. The mechanism of attention can be formulated by:
% \begin{equation}
% Attention = Softmax(A)V,\quad A = \frac{QK^{T}}{\sqrt{d_{m}}}
% \end{equation}
% where ${Q,K,V} \in \mathbb{R}^{Z\times d_{m}}$ are generated by the input feature, $Z$ is the number of tokens and $d_{m}$ is the dimension of feature. $A \in \mathbb{R}^{Z\times Z}$ denotes the attention weight matrix.
% The input of our transformer module is the noisy bone length and direction generated by the forward diffusion process. For better denoising, the 2D pose sequence is added as the condition and concatenated with the 3D noisy data as the whole input. 

\section{Experiments}\label{sec5}
\begin{table}[t]
    % \vspace{-0.5cm} % 调整表格与下方内容的间距
    \setlength{\tabcolsep}{0.3mm}
    \fontsize{9}{9}\selectfont % 字号为14pt，行距为16pt
    % \resizebox{\textwidth}{!}{%
    \renewcommand{\arraystretch}{1.0} % 将行距调整为默认的1.5倍
    
    \begin{tabular}{lcccc}
        \toprule
        \multicolumn{1}{c}{\multirow{2}{*}{Methods}} & \multirow{2}{*}{Params(M)} & \multicolumn{2}{c}{Time(min)} & \multirow{2}{*}{GFLOPs} \\
        \multicolumn{1}{c}{}                         &                          & Ref.              & Ours              &                                  \\ \midrule
        \rowcolor{gray!15}
        VideoPose3D~\cite{pavllo:videopose3d:2019}   & 8.6                     & --                  & 6.0                   & 634.7                           \\
        \rowcolor{gray!15}
        PoseFormerV1~\cite{zheng20213d}              & 9.6                     & 80.7               & 12.7                  & 661.9                           \\
        \rowcolor{gray!15}
        PoseFormerV2~\cite{Zhao_2023_CVPR}           & 14.4                    & 9.6                    & 0.8                   & 14.4                            \\ \midrule
        \rowcolor{blue!8}
        MixSTE~\cite{Zhang_2022_CVPR}                & 33.8                    & 5.1                & 1.3                   & 147.9                           \\
        \rowcolor{blue!8}
        STCFormer~\cite{tang20233d}                  & 18.9                    & --                  & 0.5                   & 26.5                            \\
        \rowcolor{blue!8}
        D3DP~\cite{shan2023diffusion}                & 34.8                    & 4.0                & 1.4                   & 147.9                           \\
        \rowcolor{blue!8}
        KTPFormer~\cite{peng2024ktpformer}           & 39.4                    & 5.5                & 1.6                   & 170.9                           \\
        \rowcolor{blue!8}
        FinePOSE~\cite{xu2024finepose}               & 200.6                   & 8.3                & 2.0                   & 156.6                           \\ \midrule
        \rowcolor{blue!8}
        DDHPose~\cite{cai2024disentangled}           & 38.5                    & 9.4                & 2.3                   & 159.2                           \\
        \rowcolor{blue!8}
        FastDDHPose                                    & 34.8                    & 4.8                  & 1.4                   & 148.0                           \\ \bottomrule
        \end{tabular}
        \caption{Training efficiency comparison under a unified framework.}
    \label{tab1}
    % \vspace{-4cm} % 调整表格与下方内容的间距
    \end{table}
\begin{table*}[t]
    % \vspace{-0.5cm} % 调整表格与下方内容的间距
    \setlength{\tabcolsep}{0.3mm}
    \fontsize{9}{9}\selectfont % 字号为14pt，行距为16pt
    % \resizebox{\textwidth}{!}{%
    \renewcommand{\arraystretch}{1.0} % 将行距调整为默认的1.5倍
    
    \begin{tabular}{lcccccccccccccccc}
        \toprule
        \multicolumn{17}{c}{\textbf{Deterministic methods: Disentangled-based model}}                                                                                                                                                                                                                                                                                                                                                                                                                           \\ \midrule
        \multicolumn{1}{l|}{\textbf{MPJPE}}  & Dir.                    & Disc.                   & Eat                     & Gre.                    & Pho.                    & Photo                   & Pos.                    & Pur.                    & Sit                     & SitD.                   & Smo.                    & Wait                    & WalkD.                  & Walk                    & \multicolumn{1}{c|}{WalkT.}                  & Avg.           \\ \midrule
        \multicolumn{1}{l|}{DKA~\cite{xu2020deep}($\mathcal{S}$)}                     & 37.4                    & 43.5                    & 42.7                    & 42.7                    & 46.6                    & 59.7                    & 41.3                    & 45.1                    & 52.7                    & 60.2                    & 45.8                    & 43.1                    & 47.7                    & 33.7                    & \multicolumn{1}{c|}{37.1}                    & 45.6                    \\
        \multicolumn{1}{l|}{Anatomy~\cite{chen2021anatomy}($\mathcal{L}$)}              & 41.4                    & 43.5                    & 40.1                    & 42.9                    & 46.6                    & 51.9                    & 41.7                    & 42.3                    & 53.9                    & 60.2                    & 45.4                    & 41.7                    & 46.0                    & 31.5                    & \multicolumn{1}{c|}{32.7}                    & 44.1                    \\
        \multicolumn{1}{l|}{Virtual Bones~\cite{wang2022motion}($\mathcal{L}$)}        & 42.4                    & 43.5                    & 41.0                    & 43.5                    & 46.7                    & 54.6                    & 42.5                    & 42.1                    & 54.9                    & 60.5                    & 45.7                    & 42.1                    & 46.5                    & 31.7                    & \multicolumn{1}{c|}{33.7}                    & 44.8                    \\ \midrule
        \rowcolor{cyan!10}
        \multicolumn{1}{l|}{DDHPose~\cite{cai2024disentangled}($\mathcal{L}$)}                      & 37.2                    & 40.3                    & 35.9                    & 38.2                    & 42.2                    & 46.8                    & 38.2                    & 37.7                    & 51.8                    & 53.3                    & 41.6                    & 39.1                    & 38.3                    & 27.3                    & \multicolumn{1}{c|}{27.9}                    & 39.7                    \\ 
        \rowcolor{cyan!10}
        \multicolumn{1}{l|}{FastDDHPose($\mathcal{L}$)}                    & 37.1                    & 39.1                    & 36.3                    & 37.1                    & 42.3                    & 48.3                    & 38.6                    & 37.8                    & 50.1                    & 53.0                    & 41.7                    & 38.5                    & 39.3                    & 27.2                    & \multicolumn{1}{c|}{27.7}                    & 39.6                    \\ \bottomrule \addlinespace[2pt] 
        \multicolumn{17}{c}{\textbf{Deterministic   methods: Non-Disentangled-based model}}                                                                                                                                                                                                                                                                                                                                                                                                                     \\ \midrule
        \multicolumn{1}{l|}{\textbf{MPJPE}}  & Dir.                    & Disc.                   & Eat                     & Gre.                    & Pho.                    & Photo                   & Pos.                    & Pur.                    & Sit                     & SitD.                   & Smo.                    & Wait                    & WalkD.                  & Walk                    & \multicolumn{1}{c|}{WalkT.}                  & Avg.           \\ \midrule
        \multicolumn{1}{l|}{VideoPose3D~\cite{pavllo:videopose3d:2019}($\mathcal{T},\dagger$)}         & 46.9 & 49.4  & 45.3 & 48.9 & 51.7 & 59.1  & 47.9 & 45.5 & 58.4 & 65.9  & 50.3 & 47.1 & 53.2   & 37.2 & \multicolumn{1}{c|}{39.9}   & 49.8                    \\
        \multicolumn{1}{l|}{PoseFormerV1~\cite{zheng20213d}($\mathcal{T},\dagger$)}        & 49.8 & 53.1  & 47.0 & 52.2 & 52.9 & 59.9  & 50.6 & 47.8 & 60.1 & 69.3  & 53.1 & 49.4 & 53.0   & 37.0 & \multicolumn{1}{c|}{40.5}   & 51.7          \\
        \multicolumn{1}{l|}{P-STMO~\cite{shan2022p}($\mathcal{L}$)}      & 38.9          & 42.7          & 40.4          & 41.1          & 45.6          & 49.7          & 40.9          & 39.9          & 55.5          & 59.4          & 44.9          & 42.2          & 42.7          & 29.4          & \multicolumn{1}{c|}{29.4}          & 42.8          \\
        \multicolumn{1}{l|}{MixSTE~\cite{Zhang_2022_CVPR}($\mathcal{L},\dagger$)}        & 38.6 & 40.0  & 36.6 & 39.1 & 43.1 & 50.2  & 38.5 & 39.1 & 52.7 & 57.5  & 42.4 & 40.2 & 40.4   & 28.2 & \multicolumn{1}{c|}{28.8}   & 41.0          \\
        \multicolumn{1}{l|}{PoseFormerV2~\cite{Zhao_2023_CVPR}($\mathcal{T},\dagger$)}        & 47.4 & 50.1  & 45.1 & 49.5 & 51.8 & 56.9  & 47.5 & 46.2 & 60.2 & 68.4  & 50.0 & 47.0 & 52.3   & 36.1 & \multicolumn{1}{c|}{37.9}   & 49.8          \\
        \multicolumn{1}{l|}{STCFormer~\cite{tang20233d}($\mathcal{L},\dagger$)}        & 39.5 & 42.5  & 39.8 & 39.9 & 43.6 & 53.8  & 41.8 & 39.7 & 54.1 & 59.4  & 43.8 & 42.8 & 42.4   & 28.9 & \multicolumn{1}{c|}{29.3}   & 42.7          \\
        % \multicolumn{1}{l|}{D3DP~\cite{shan2023diffusion}($\mathcal{L},\dagger$)}        & 37.1                 & 40.4                 & 36.0                 & 38.4                 & 42.3                 & 49.2                 & 38.3                 & 38.9                 & 50.3                 & 54.8                 & 41.6                 & 39.1                 & 40.1                 & 27.2                 & \multicolumn{1}{c|}{27.4}                 & 40.0                 \\
        % \multicolumn{1}{l|}{FinePose~\cite{xu2024finepose}($\mathcal{L},\dagger$)}        & 38.2                 & 40.6                 & 35.4                 & 37.9                 & 43.1                 & 50.7                 & 38.7                 & 38.8                 & 51.1                 & 56.0                 & 42.7                 & 40.6                 & 40.5                 & 27.6                 & \multicolumn{1}{c|}{27.9}                 & 40.7                 \\
        \multicolumn{1}{l|}{KTPFormer~\cite{peng2024ktpformer}($\mathcal{L},\dagger$)}                  & 37.1                    & 40.3                    & 36.8                    & 37.9                    & 42.9                    & 51.5                    & 40.5                    & 38.8                    & 51.7                    & 56.5                    & 43.1                    & 40.6                    & 40.9                    & 28.5                    & \multicolumn{1}{c|}{28.6}                    & 41.0                    \\ 
        \multicolumn{1}{l|}{HiPART~\cite{zheng2025hipart}($\mathcal{L}$)}                       & 42.8                    & 42.7                    & 38.1                    & 41.3                    & 42.7                    & 46.3                    & 37.2                    & 44.2                    & 51.0                    & 51.4                    & 40.9                    & 38.3                    & 40.0                    & 39.9                    & \multicolumn{1}{c|}{33.7}                    & 42.0                    \\        \midrule
        \rowcolor{cyan!10}
        \multicolumn{1}{l|}{DDHPose~\cite{cai2024disentangled}($\mathcal{L}$)}               & 37.2                    & 40.3                    & 35.9                    & 38.2                    & 42.2                    & 46.8                    & 38.2                    & 37.7                    & 51.8                    & 53.3                    & 41.6                    & 39.1                    & 38.3                    & 27.3                    & \multicolumn{1}{c|}{27.9}                    & 39.7                    \\
        \rowcolor{cyan!10}
        \multicolumn{1}{l|}{FastDDHPose($\mathcal{L}$)}                    & 37.1                    & 39.1                    & 36.3                    & 37.1                    & 42.3                    & 48.3                    & 38.6                    & 37.8                    & 50.1                    & 53.0                    & 41.7                    & 38.5                    & 39.3                    & 27.2                    & \multicolumn{1}{c|}{27.7}                    & 39.6                    \\ \bottomrule \addlinespace[2pt] 
        \multicolumn{17}{c}{\textbf{Probabilistic   methods}}                                                                                                                                                                                                                                                                                                                                                                                                                                                   \\ \midrule
        \multicolumn{1}{l|}{\textbf{MPJPE}}  & Dir.                    & Disc.                   & Eat                     & Gre.                    & Pho.                    & Photo                   & Pos.                    & Pur.                    & Sit                     & SitD.                   & Smo.                    & Wait                    & WalkD.                  & Walk                    & \multicolumn{1}{c|}{WalkT.}                  & Avg.           \\ \midrule
        \multicolumn{1}{l|}{MHFormer~\cite{li2022mhformer}($\mathcal{X}$,$H{=}3$)} & 39.2                    & 43.1                    & 40.1                    & 40.9                    & 44.9                    & 51.2                    & 40.6                    & 41.3                    & 53.5                    & 60.3                    & 43.7                    & 41.1                    & 43.8                    & 29.8                    & \multicolumn{1}{c|}{30.6}                    & 43.0                    \\
        \multicolumn{1}{l|}{GFPose~\cite{ci2023gfpose}($H{=}10$)}              & 39.9                    & 44.6                    & 40.2                    & 41.3                    & 46.7                    & 53.6                    & 41.9                    & 40.4                    & 52.1                    & 67.1                    & 45.7                    & 42.9                    & 46.1                    & 36.5                    & \multicolumn{1}{c|}{38.0}                    & 45.1                    \\
        \multicolumn{1}{l|}{D3DP~\cite{shan2023diffusion}($\mathcal{L},\dagger$,$H{=}20$)}    & 37.9 & 39.7  & 36.0 & 37.6 & 42.0 & 48.2  & 38.6 & 38.4 & 49.8 & 54.7  & 41.7 & 38.8 & 39.6   & 27.1 & \multicolumn{1}{c|}{27.4}   & 39.8          \\
        \multicolumn{1}{l|}{FinePose~\cite{xu2024finepose}($\mathcal{L},\dagger$,$H{=}20$)}                   & 38.1 & 40.3  & 35.2 & 37.7 & 42.8 & 50.4  & 38.5 & 38.5 & 50.7 & 55.3  & 42.4 & 40.2 & 40.1   & 27.4 & \multicolumn{1}{c|}{27.7}   & 40.4          \\ \midrule
        \rowcolor{cyan!10}
        \multicolumn{1}{l|}{DDHPose~\cite{cai2024disentangled}($\mathcal{L}$,$H{=}20$)}   & 36.4                    & 39.5                    & 34.9                    & 37.6                    & 40.1                    & 45.9                    & 37.8                    & 37.8                    & 51.5                    & 52.2                    & 40.8                    & 38.3                    & 38.3                    & 27.0                    & \multicolumn{1}{c|}{27.0}                    & 39.0                    \\
        \rowcolor{cyan!10}
        \multicolumn{1}{l|}{FastDDHPose($\mathcal{L}$,$H{=}20$)} & 36.6 & 38.7  & 35.9 & 36.8 & 42.0 & 47.7  & 38.1 & 37.3 & 49.8 & 52.0  & 41.3 & 38.1 & 38.9   & 26.9 & \multicolumn{1}{c|}{27.2}   & 39.1          \\ \bottomrule \addlinespace[2pt]
        \end{tabular}
    
    % }
    % \vspace{-0.5cm}
    \caption{Results on Human3.6M in millimeters under MPJPE. 
    $\mathcal{S}$, $\mathcal{T}$, $\mathcal{L}$, and $\mathcal{X}$ represent models with receptive fields of 9, 27, 243, and 351 input frames, respectively. 
    $H$ denotes the number of hypothesis poses in probabilistic methods. 
    $\dagger$ indicates results reproduced by us using the Fast3DHPE framework. 
    For brevity, the iteration number $W{=}10$ used in the inference stage of probabilistic methods D3DP, FinePose, DDHPose, and FastDDHPose is omitted.}
    
    \label{tab2}
    % \vspace{-4cm} % 调整表格与下方内容的间距
    \end{table*}

In this section, we present a comprehensive evaluation of the unified framework Fast3DHPE and our proposed FastDDHPose. All models in this paper including the reproduced mainstream baselines and FastDDHPose, are trained and evaluated on eight NVIDIA GeForce RTX 3090 GPUs within the Fast3DHPE framework. 

\subsection{Efficiency Comparison}
We first evaluate the efficiency gains introduced by the Fast3DHPE framework through a unified comparison with other state-of-the-art 3D human pose estimation methods. We then conduct a detailed efficiency comparison between FastDDHPose and DDHPose, highlighting the improvements brought by the proposed KHSTDenoiser. The corresponding observations are summarized as follows:

\begin{enumerate}
    \item \textbf{Parameter Size.}  
    Methods highlighted with a gray background in Table~\ref{tab1} correspond to the \textit{seq2frame} paradigm, while those highlighted in blue represent \textit{seq2seq} approaches. Seq2frame models typically contain fewer parameters because they capture local temporal cues and do not construct global temporal dependencies, whereas seq2seq methods require heavier temporal modeling modules to learn long-range relationships across the entire receptive field.
    \footnotetext[3]{PyTorch DataParallel documentation: \url{https://pytorch.org/docs/stable/generated/torch.nn.DataParallel.html}}

    \item \textbf{Training Efficiency.}  
    We compare the training time per epoch reported in the original papers (\textit{DP}-based training\footnotemark{}) with the performance under our unified Fast3DHPE framework. With the integration of \textit{DDP} and \textit{AMP}, all models achieve a remarkable 3--10 times speedup in training time, significantly improving iteration efficiency and enabling more practical large-scale experimentation.

    \item \textbf{Computational Complexity.}  
    We evaluate the computational cost by measuring the GFLOPs required to generate 243-frame predictions with receptive field of 243. \textit{Seq2frame} methods incur significantly higher computation because the model processes a full sequence but outputs only a single frame at a time. Producing 243 frames therefore requires 243 forward passes, leading to substantial redundant computation. In contrast, \textit{seq2seq} methods generate the entire sequence in a single forward pass and are thus far more efficient. Notably, PoseFormerV2 further reduces the computational burden by performing temporal modeling in the frequency domain.

    \item \textbf{Efficiency Improvements of FastDDHPose.}  
    From the method perspective, FastDDHPose significantly reduces training time compared with DDHPose, achieving a \textbf{48.9\%} reduction in per-epoch training time under the original framework (from 9.4 to 4.8 minutes), and a further \textbf{39.1\%} reduction under the Fast3DHPE framework (from 2.3 to 1.4 minutes).
    By jointly leveraging method-level optimization and the unified Fast3DHPE framework, the per-epoch training time is reduced by approximately \textbf{85.1\%}, from 9.4 to 1.4 minutes. Meanwhile, FastDDHPose reduces the parameter size by \textbf{9.6\%} and GFLOPs by \textbf{7\%}, while achieving slightly better performance on Human3.6M compared with DDHPose (MPJPE 39.6 mm vs. 39.7 mm).
\end{enumerate}

\subsection{Quantitative Results}
\subsubsection{Results on Human3.6M}
The results of our method on Human3.6M are presented in Table~\ref{tab2}. We begin by comparing our approach with state-of-the-art deterministic 3D human pose estimation methods. Deterministic methods evaluate performance based solely on a single forward prediction of the model without any iterative refinement, and therefore provide a clearer indication of how the model behaves in real-world deployment scenarios.

\begin{table}[t]
    % \vspace{-0.5cm} % 调整表格与下方内容的间距
    \setlength{\tabcolsep}{0.4mm}
    \fontsize{9}{9}\selectfont % 字号为14pt，行距为16pt
    % \resizebox{\textwidth}{!}{%
    \renewcommand{\arraystretch}{1.0} % 将行距调整为默认的1.5倍
    
    \begin{tabular}{l|cc}
        \toprule
        \multicolumn{1}{c|}{Methods}                  & MPJPE & P-MPJPE \\ \midrule
        DiffPose~\cite{gong2023diffpose}($H$=5, $W$=50)            & 36.9  & 28.7    \\
        DiffPose$\sharp$~\cite{gong2023diffpose}($H$=5, $W$=50)            & 40.1  & 31.1    \\ \midrule
        FastDDHPose ($H$=5, $W$=50) & 39.3  & 31.1    \\ \bottomrule
        \end{tabular}
        \caption{Comparison with DiffPose on Human3.6M. ($\sharp$)- Stand-Diff implemented in DiffPose.}
    \label{tab3}
    % \vspace{-4cm} % 调整表格与下方内容的间距
    \end{table}

To facilitate a structured comparison, we categorize existing methods into two groups depending on whether the regression of 3D joint locations is decomposed into bone-length and bone-direction components: (i) disentangle-based methods, and (ii) non-disentangle-based methods. For disentangle-based methods, we can see from the table that our method achieves the best MPJPE of 39.6mm, surpassing Anatomy3D~\cite{chen2021anatomy} by 4.5mm(10.2\%) in MPJPE. For non-disentangle based model, we improve KTPFormer~\cite{peng2024ktpformer} by 1.4mm(3.4\%) under MPJPE. And then we compare our method with probabilistic methods, our method reaches the SOTA MPJPE of 39.0mm, outerperforms  D3DP ~\cite{shan2023diffusion} by 0.8mm(2.0\%). 

As for DiffPose~\cite{gong2023diffpose}, we separately compare with it in Table~\ref{tab3}. Note that, the DiffPose additionally introduces the heatmaps derived from an off-the-shelf 2D pose detector and depth distributions to initialize the pose distribution. The probabilistic methods in Table~\ref{tab2} only use the 2D pose sequences. Thus, it might not be fair to directly compare with DiffPose. But according to DiffPose, the implementation of Stand-Diff only uses 2D pose sequences by reversing the 3D pose from a standard Gaussian noise, which achieves a larger MPJPE error than our FastDDHPose with the same setting (40.1mm vs 39.3mm). The results demonstrate that our method can notably improves performance by 0.8mm through the Disentangle Strategy and the utilization of hierarchical relations.

\subsubsection{Results on MPI-INF-3DHP}
We also evaluate our method on the MPI-INF-3DHP dataset under PCK, AUC, and MPJPE metrics. In Table~\ref{tab4}, our approach outperforms the SOTA method by 0.8 in PCK, 0.4 in AUC, and 1.2mm in MPJPE under the single hypothesis condition.
\begin{table}[t]
    % \vspace{-0.5cm} % 调整表格与下方内容的间距
    \setlength{\tabcolsep}{0.4mm}
    \fontsize{9}{9}\selectfont % 字号为14pt，行距为16pt
    % \resizebox{\textwidth}{!}{%
    \renewcommand{\arraystretch}{1.1} % 将行距调整为默认的1.5倍
    
    \begin{tabular}{l|ccc}
        \toprule
        \multicolumn{1}{c|}{Methods}       & PCK$\uparrow$  & AUC$\uparrow$  & MPJPE$\downarrow$ \\ \midrule
        Anatomy3D~\cite{chen2021anatomy}($\mathcal{L}$)            & 87.8 & 53.8 & 79.1  \\
        PoseFormerV1~\cite{zheng20213d}($\mathcal{S}$)            & 88.6 & 56.4 & 77.1  \\
        P-STMO~\cite{shan2022p}($\mathcal{M}$)            & 97.9 & 75.8 & 32.2  \\
        MixSTE~\cite{Zhang_2022_CVPR}($\mathcal{L}$)           & 96.9 & 75.8 & 35.4  \\
        D3DP~\cite{shan2023diffusion}($\mathcal{L}$,$H{=}1$,$W{=}1$)        & 97.7 & 77.8 & 30.2  \\ \midrule
        DDHPose~\cite{cai2024disentangled} ($\mathcal{L}$,$H{=}1$,$W{=}1$) & \textbf{98.5} & 78.1 & \textbf{29.2}  \\ 
        FastDDHPose($\mathcal{L}$,$H{=}1$,$W{=}1$) & 98.3 & \textbf{78.2} & \textbf{29.2}  \\\bottomrule
        \end{tabular}
    
    % }
    % \vspace{-0.5cm}
    \caption{Results on MPI-INF-3DHP under PCK, AUC, and MPJPE using ground truth 2D pose as inputs. The best results are highlighted in bold.}
    
    \label{tab4}
    % \vspace{-4cm} % 调整表格与下方内容的间距
    \end{table}

\subsubsection{Ablation Study}
In order to evaluate each design in our method, we conduct ablation experiments on the Human3.6M dataset using 2D pose sequence extracted by CPN. 

\begin{table}[t]
    % \vspace{-0.5cm} % 调整表格与下方内容的间距
    \setlength{\tabcolsep}{1.0mm}
    \fontsize{9}{9}\selectfont % 字号为14pt，行距为16pt
    % \resizebox{\textwidth}{!}{%
    \renewcommand{\arraystretch}{1.0} % 将行距调整为默认的1.5倍
    
    \begin{tabular}{cc|cc}
        \toprule
        Dis. Input & Dis. Output & MPJPE & P-MPJPE \\ \midrule
        $\times$                  & $\times$             & 40.0  & 31.6   \\
        $\times$                  & $\surd$              & 42.0   & 33.4        \\
        $\surd$             & $\times$                   & \textbf{39.6} & \textbf{31.3}   \\
        $\surd$              & $\surd$                  & 40.5 & 32.1   \\ \bottomrule
        \end{tabular}
        \caption{The impact of disentanglement strategy. The disentanglement strategy with Disentangled input and without Disentangled output has the best result highlighted in bold.}
    \label{tab5}
    % \vspace{-4cm} % 调整表格与下方内容的间距
    \end{table}
\begin{figure}[t]
    \centering
    \includegraphics[width=1\columnwidth]{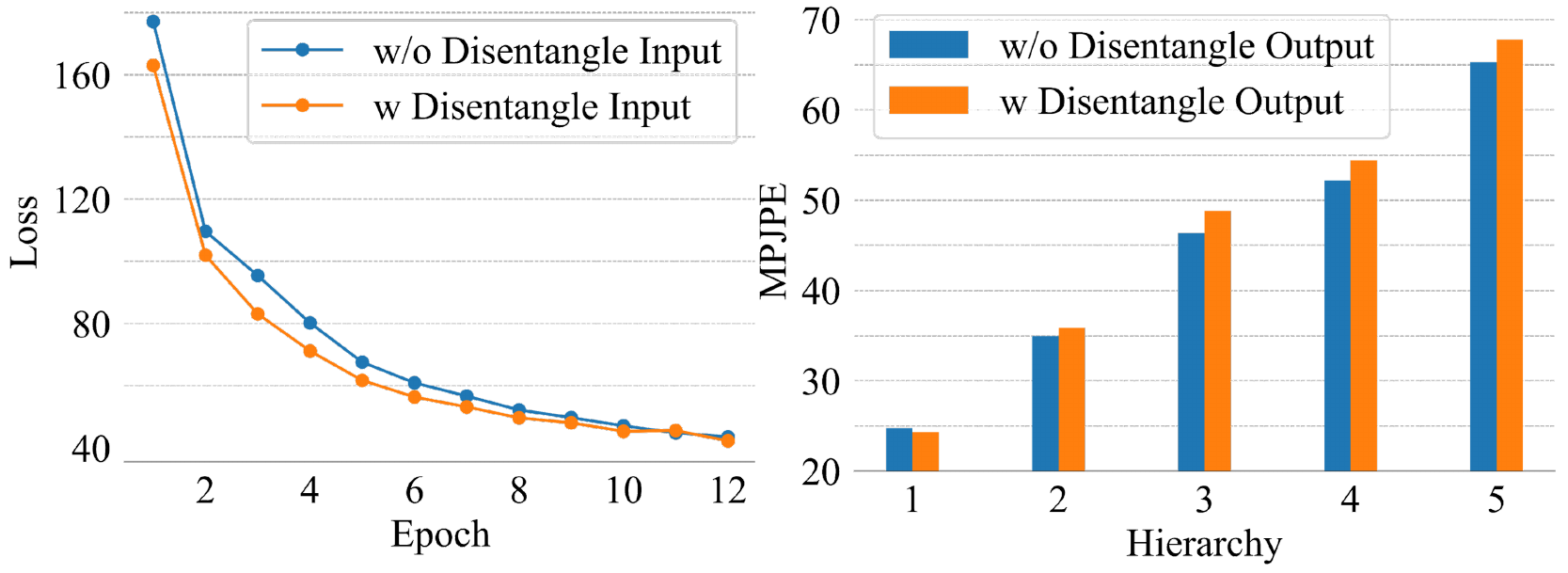} % Reduce the figure size so that it is slightly narrower than the column. Don't use precise values for figure width.This setup will avoid overfull boxes.
    \caption{Left: Training Loss Comparison (w/o Disentangle Output). Right: Hierarchical Error Comparison (w/o Disentangle Input).}
    % \vspace{-3.85ex}
    \label{fig7}
    \end{figure}
\begin{figure*}[t]
    \centering
    \includegraphics[width=\textwidth]{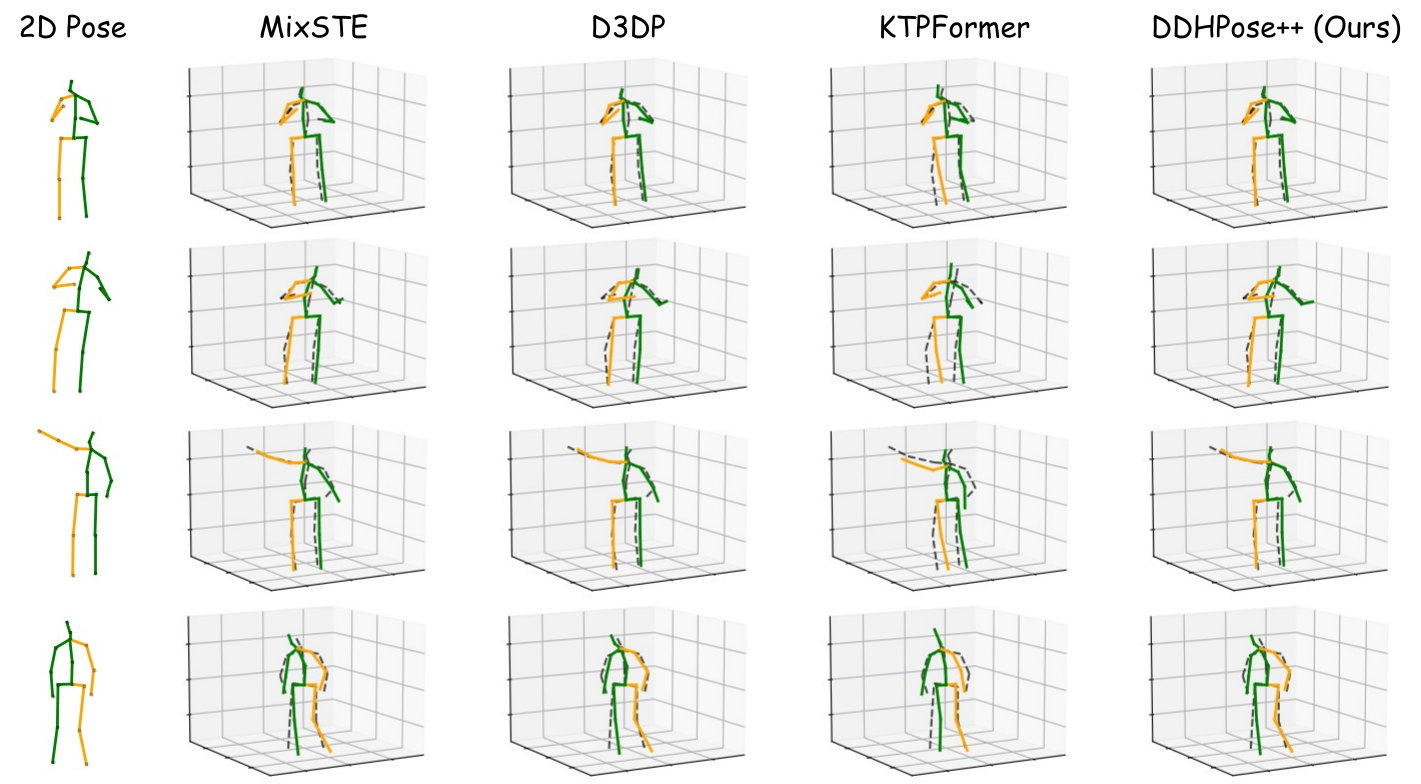} % Reduce the figure size so that it is slightly narrower than the column. Don't use precise values for figure width.This setup will avoid overfull boxes.
    \caption{Qualitative comparison on Human3.6M among our method, MixSTE~\cite{Zhang_2022_CVPR}, D3DP~\cite{shan2023diffusion}, and KTPFormer~\cite{peng2024ktpformer}. Colored solid lines denote the predicted 3D poses, while gray dashed lines represent the ground-truth poses.}
    \label{fig6}
\end{figure*}

\paragraph{Disentanglement Strategy}
In this section, we separately compare the effect of the Disentangle Input and Disentangle Output strategy. 
\par For the Disentangle Input Strategy, our method divides the dense and high-dimensional optimization problem into two low-dimensional sub-problems, simplifying the learning of the human pose prior. As shown in the left portion of Fig.~\ref{fig7}, employing the Disentangle Input strategy results in faster convergence and lower training 3D pose loss compared to not using it in the initial training epoch. This leads to improved quantitative results (39.6mm vs 40.0mm), as highlighted in Table~\ref{tab5}.

\par For Disentangle Output, the denoiser in the reverse process directly regresses bone length and direction, generating the 3D pose using $C = C_{p}+l \times d$, where $C$ and $C_{p}$ are joint and parent joint coordinates, and $l$, $d$ represent predicted bone length and direction. This equation indicates that a joint's coordinate depends not only on its own bone properties but also on all parent joints along the bone chain. As illustrated in the right portion of Fig.~\ref{fig7}, hierarchy 1 exhibits lower errors in the Disentangled Output setting, while higher hierarchical levels accumulate errors more than without using Disentangled Output. Quantitative results in Table~\ref{tab5} show that employing the Disentangle Output strategy increases MPJPE from 40.0mm to 42.0mm.

\begin{figure*}[t]
    \centering
    \includegraphics[width=\textwidth]{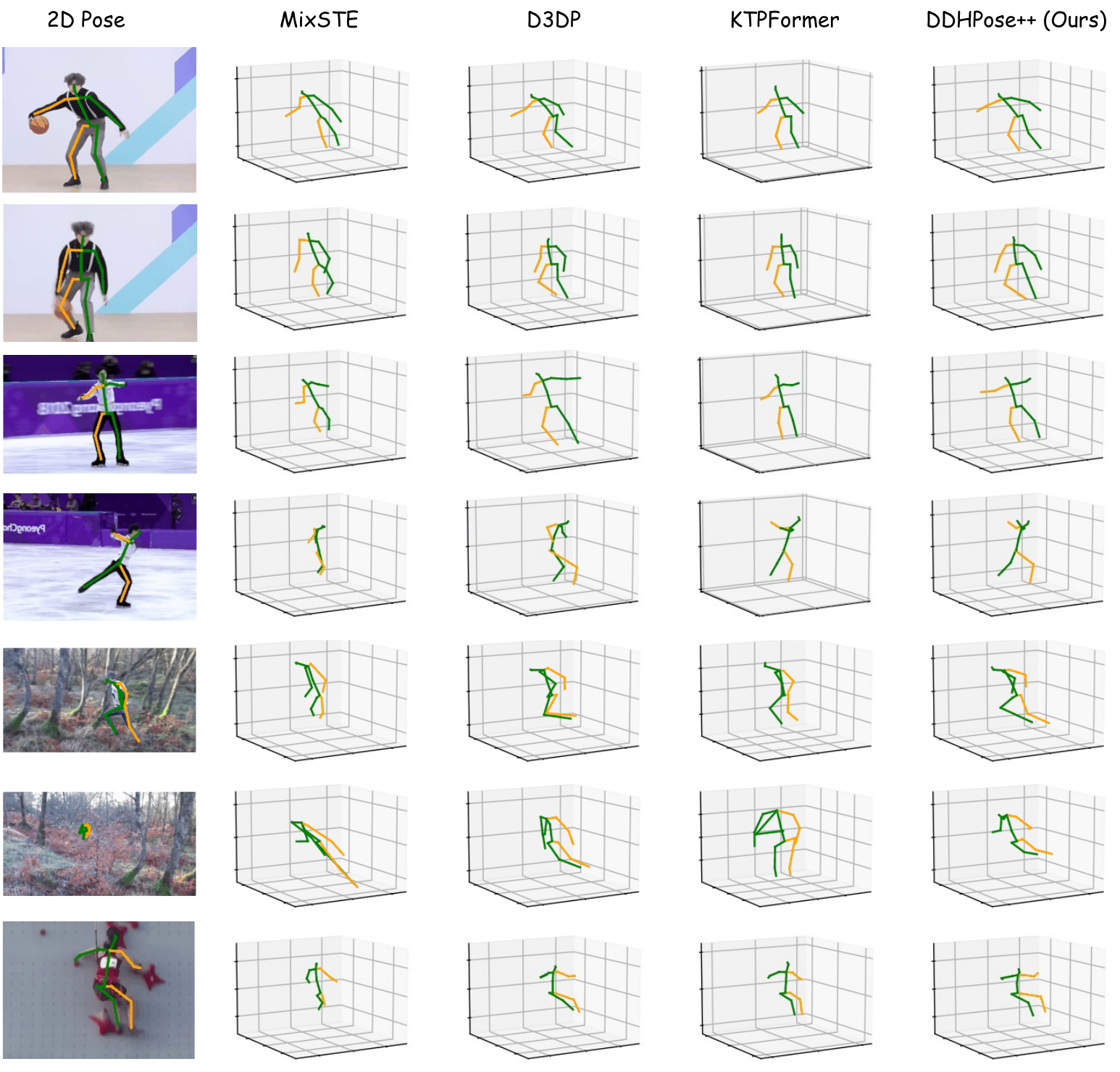} % Reduce the figure size so that it is slightly narrower than the column. Don't use precise values for figure width.This setup will avoid overfull boxes.
    \caption{Qualitative results comparing our method with MixSTE~\cite{Zhang_2022_CVPR}, D3DP~\cite{shan2023diffusion}, and KTPFormer~\cite{peng2024ktpformer} on in-the-wild videos from YouTube. For preprocessing, person detection is performed using Faster R-CNN~\cite{girshick2015fast}, and 2D keypoints are estimated using HRNet~\cite{wang2020deep} integrated in MMPose~\cite{mmpose2020}.}
    \label{fig8}
\end{figure*}

\paragraph{Effect of each module}
As summarized in Table~\ref{tab6}, our framework consists of three major components: the Disentangle Strategy (Disentangle), the Hierarchical Embedding module (HiE), and the Kinematic–Hierarchical Spatial Transformer (KHST). In these experiments, we fix both $H$ and $W$ to 1 to ensure fair comparison.

The result shows that \emph{Disentangle} slightly amplifies the accumulated error. Adding \emph{HiE} improves MPJPE from 40.3mm to 39.9mm and lifts P-MPJPE from 31.8mm to 31.5mm. Further integrating \emph{KHST} refines the MPJPE from 39.9mm to 39.6mm and improves P-MPJPE from 31.5mm to 31.3mm. 

The results demonstrate that the hierarchical relationships among different joints have a substantial impact on the model’s performance. In particular, assigning different levels of attention to different hierarchy levels, with stronger constraints on higher level joints, proves to be especially beneficial.

\begin{table}[t]
    % \vspace{-0.5cm} % 调整表格与下方内容的间距
    \setlength{\tabcolsep}{1.0mm}
    \fontsize{9}{9}\selectfont % 字号为14pt，行距为16pt
    % \resizebox{\textwidth}{!}{%
    \renewcommand{\arraystretch}{1.0} % 将行距调整为默认的1.5倍

        \begin{tabular}{ccc|cc}
            \toprule
            Disentangle & HiE & KHST & \multicolumn{1}{c}{MPJPE} & \multicolumn{1}{c}{P-MPJPE} \\ \midrule
            $\times$    & $\times$   & $\times$    & 40.0                      & 31.8                        \\
            $\surd$     & $\times$   & $\times$    & 40.3                      & 31.8                        \\
            $\surd$     & $\surd$   & $\times$     & 39.9                      & 31.5                            \\
            $\surd$     & $\surd$   & $\surd$    & \textbf{39.6}                      & \textbf{31.3}                        \\ \bottomrule
            \end{tabular}
            \caption{The impact of disentanglement strategy. The disentanglement strategy with Disentangled input and without Disentangled output has the best result highlighted in bold.}
    \label{tab6}
    % \vspace{-4cm} % 调整表格与下方内容的间距
    \end{table}
\begin{table}[t]
    % \vspace{-0.5cm} % 调整表格与下方内容的间距
    \setlength{\tabcolsep}{1.0mm}
    \fontsize{9}{9}\selectfont % 字号为14pt，行距为16pt
    % \resizebox{\textwidth}{!}{%
    \renewcommand{\arraystretch}{1.1} % 将行距调整为默认的1.5倍
    
    \begin{tabular}{cccc}
        \toprule
        3D Pose   Loss & 3D Dis. Loss & MPJPE & P-MPJPE \\ \midrule
        $\surd$              & $\times$                  & 40.5  & 32.2    \\
        $\surd$              & $\surd$                  & \textbf{39.6}  & \textbf{31.3}    \\ \bottomrule
        \end{tabular}
        \caption{Ablation study for loss function proposed in our method. The best results are highlighted in bold.}
    \label{tab7}
    % \vspace{-4cm} % 调整表格与下方内容的间距
    \end{table}

\paragraph{Effect of Loss Function}
We employ the 3D pose loss and 3D disentanglement loss to train FastDDHPose.
3D pose loss is used to constrain the denoised 3D pose regressed by our model and the 3D disentanglement loss is utilized to aid the model in learning the explicit human body prior during the forward diffusion process. The contribution of the loss function is in Table~\ref{tab7}. The result shows that using 3D disentanglement loss is essential for a better result, improving MPJPE from 40.3mm to 39.9mm and lifting P-MPJPE from 31.8mm to 31.5mm. Moreover, the 3D disentanglement loss serves as an effective plug-and-play loss function that can be readily applied to existing 3D HPE pipelines. We have also integrated a disentanglement-loss interface into Fast3DHPE to facilitate its adoption.

\subsection{Qualitative Results}
We further provide qualitative comparisons on both Human3.6M and in-the-wild scenarios to evaluate the visual plausibility of our predictions. As shown in Fig.~\ref{fig6}, FastDDHPose produces stable and coherent 3D poses on Human3.6M, with notably accurate estimations on high-level joints such as the legs and upper limbs. These results indicate that the hierarchical modeling and disentanglement strategy effectively enhance the structural reliability of the predicted poses.

In more challenging in-the-wild scenes in Fig.~\ref{fig8}, our method is able to reconstruct the underlying motion with high fidelity while preserving realistic human-body proportions under the guidance of explicit human priors. Even when the input 2D pose contains small-scale or low-resolution joint coordinates, FastDDHPose remains the most capable model in recovering a plausible full-body structure, demonstrating strong robustness and generalization across diverse visual conditions.

\section{Conclusion}\label{sec6}
We propose FastDDHPose, a disentangled diffusion-based framework for 3D human pose estimation that incorporates hierarchical information in both the forward diffusion and reverse denoising processes. By disentangling bone length and direction based on the kinematic hierarchy and introducing the KHSTDenoiser to enhance hierarchical joint relationships, FastDDHPose achieves more stable performance and consistently outperforms existing disentangle-based, non-disentangle-based, and probabilistic methods on Human3.6M and MPI-INF-3DHP.

We further build Fast3DHPE, a unified and standardized framework that integrates mainstream 3D HPE methods under a consistent and configurable pipeline. Fast3DHPE supports efficient multi-GPU training and systematic evaluation, enabling fair comparison and reproducible benchmarking. Within this framework, FastDDHPose demonstrates stable behavior in qualitative in-the-wild visualizations.

Future research will focus on improving robustness and cross-domain generalization of 3D HPE models in real-world scenarios, addressing challenges posed by diverse environments and dataset shifts. We hope FastDDHPose and Fast3DHPE can facilitate future advances toward practical and generalizable 3D human pose estimation.

\section{Statements and Declarations}\label{sec7}
\subsection{Competing Interests}
The authors declare that they have no known competing financial or non-financial interests that could have appeared to influence the work reported in this paper.

\subsection{Data Availability}
The data used in this study are publicly available benchmark datasets, and no new datasets were generated during this work.

\begin{appendices}

% \section{Section title of first appendix}\label{secA1}

% An appendix contains supplementary information that is not an essential part of the text itself but which may be helpful in providing a more comprehensive understanding of the research problem or it is information that is too cumbersome to be included in the body of the paper.

\end{appendices}


\bibliography{sn-bibliography}% common bib file

\end{document}